\pdfoutput=1
\documentclass[sigconf]{acmart}

\usepackage{multirow}
\usepackage[caption=false]{subfig}
\usepackage{algorithm}
\usepackage{amsfonts}
\usepackage{booktabs}
\usepackage{float}
\usepackage{wrapfig}
\usepackage{amsmath}
\usepackage{diagbox}
\usepackage{balance}
\usepackage{algpseudocode}
\newcommand{\colorr}[1]{\textcolor{black}{#1}}
\newcommand{\colorb}[1]{\textcolor{black}{#1}}
\usepackage{xcolor}         
 
\usepackage{amssymb}

\AtBeginDocument{%
  \providecommand\BibTeX{{%
    \normalfont B\kern-0.5em{\scshape i\kern-0.25em b}\kern-0.8em\TeX}}}

\setcopyright{acmlicensed}
\copyrightyear{2018}
\acmYear{2018}
\acmDOI{XXXXXXX.XXXXXXX}

\acmConference[Conference acronym 'XX]{Make sure to enter the correct
  conference title from your rights confirmation emai}{June 03--05,
  2018}{Woodstock, NY}
\acmISBN{978-1-4503-XXXX-X/18/06}

\begin{document}

\title{Diversity Optimization for Travelling Salesman Problem via Deep Reinforcement Learning}

\author{Qi Li}
\affiliation{%
  \institution{South China University of Technology}
  \city{Guangzhou}
  \country{China}
}
\email{researchliqi@gmail.com}

\author{Zhiguang Cao}
\affiliation{%
  \institution{Singapore Management University}
  \country{Singapore} 
}
\email{zhiguangcao@outlook.com}

\author{Yining Ma}
\affiliation{%
  \institution{Massachusetts Institute of Technology}
  \country{Cambridge, MA, United States}
}
\email{yiningma@mit.edu}

\author{Yaoxin Wu}
\affiliation{%
  \institution{Eindhoven University of Technology}
  \city{Eindhoven}
  \country{Netherlands}
}
\email{wyxacc@hotmail.com}

\author{Yue-Jiao~Gong}
\authornote{Yue-Jiao~Gong is the corresponding author.}
\affiliation{%
  \institution{South China University of Technology}
  \city{Guangzhou}
  \country{China}
}
\email{gongyuejiao@gmail.com}

\renewcommand{\shortauthors}{Li et al.}

\begin{abstract}

Existing neural methods for the Travelling Salesman Problem (TSP) mostly aim at finding a single optimal solution. To discover diverse yet high-quality solutions for Multi-Solution TSP (MSTSP), we propose a novel deep reinforcement learning based neural solver, which is primarily featured by an encoder-decoder structured policy. Concretely, on the one hand, a Relativization Filter (RF) is designed to enhance the robustness of the encoder to affine transformations of the instances, so as to potentially improve the quality of the found solutions. On the other hand, a Multi-Attentive Adaptive Active Search (MA3S) is tailored to allow the decoders to strike a balance between the optimality and diversity. Experimental evaluations on benchmark instances demonstrate the superiority of our method over recent neural baselines across different metrics, and its competitive performance against state-of-the-art traditional heuristics with significantly reduced computational time, ranging from $1.3\times$ to $15\times$ faster. Furthermore, we demonstrate that our method can also be applied to the Capacitated Vehicle Routing Problem (CVRP).

\end{abstract}





\maketitle
\section{Introduction}\label{Intro}

\begin{figure}[h]
    \centering\includegraphics[width=1.0\columnwidth]{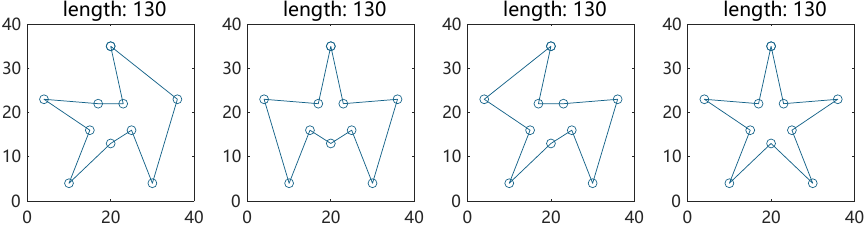}
    \caption{Illustration of a TSP-10 instance with multiple optima of equal length (retrieved from~\cite{huang2018seeking}).}
    \label{MSTSP8}
\end{figure}

The Travelling Salesman Problem (TSP), as a classic optimization problem, involves a salesman who embarks from a city
, visits each remaining city once, and finally returns to the starting one, aiming to find an optimal route with the shortest length while satisfying the above constraints~\cite{ 10.1145/3219819.3220111, 10.1145/3394486.3403355, kim2021learning, 10.1145/3534678.3539296, 10.1145/2939672.2939862, 10.1145/3580305.3599425, 10.1145/3534678.3539037, NEURIPS2023_9bae70d3, xin2020step, zhang2022learning2}. 
Nevertheless, recent studies have shown that diverse landscapes are common in many practical TSP instances, yielding the Multi-Solution TSP (MSTSP, a.k.a. the diversity optimization for TSP)~\cite{ronald1995finding, angus2006niching, han2018multimodal, huang2018seeking, huang2019niching, do2022niching}. For instance, Figure~\ref{MSTSP8} showcases a TSP-10 instance with as many as 56 optimal solutions (we only display 4 of them for illustration), each possessing an identical length of 130. Accordingly, MSTSP is in pursuit of \emph{a set of} diverse yet high-quality (possibly optimal) solutions. As a practical and crucial supplement to the classic TSP, it is highly desired in many real-world scenarios, where a single solution may be insufficient. For example, 1) when the single target route (solution) becomes unavailable due to unexpected circumstances, MSTSP offers desirable alternatives; 2) while the single target route may overlook other important metrics like user preferences, MSTSP allows for personalized choices among a set of high-quality candidate routes; 3) while the single target route may incur spontaneous and simultaneous pursuit of the same choice, MSTSP can distribute users or loads across different routes, potentially mitigating the jam and enhancing the overall performance.

Given that the classic TSP is already NP-hard, solving the MSTSP is intrinsically more tough. Thereby, the mainstream methods for MSTSP always leverage traditional heuristics~\cite{ronald1995finding, angus2006niching, han2018multimodal, huang2018seeking, huang2019niching, do2022niching}. Compared with the exact methods, they could attain good solutions more efficiently. However, the computation time still grows rapidly as the problems scale up due to the NP-hard and multi-solution nature. Moreover, the traditional heuristics always rely on hand-crafted rules and domain knowledge, which ignore the underlying pattern among the problem instances and thus likely hold back the performance especially in terms of computation efficiency.

Recently, neural heuristics have garnered considerable attention as promising alternatives to solve combinatorial optimization problems (COPs)~\cite{bengio2021machine,MAZYAVKINA2021105400,tong2021combinatorial,zhang2023review,zhou2023learning} in a data-driven fashion. 
Benefiting from the deep (reinforcement) learning and the large set of training instances, they could automatically discover decision-making rules by leveraging the transferable pattern among instances, so as to speed up the per-instance computation while improving the solution quality. Despite much success achieved, most neural heuristics emphasize a single (optimal) solution. While a number of them have the potential to deliver multiple solutions for TSP, through either multiple starting points~\cite{kwon2020pomo,kimsym,hottung2022efficient} or multiple decoders~\cite{xin2021multi}, they are still far from satisfactory. On the one hand, in spite of the symmetry~\cite{kwon2020pomo,kimsym} used for augmentation, the solutions obtained still exhibit a high similarity or low optimality without explicitly considering more affine transformations among the instances. On the other hand, although multiple decoders are used to yield different solutions~\cite{xin2021multi},
the diversity and optimality of those obtained solutions are limited without explicitly balancing them.

To tackle these challenges, we propose the first neural heuristic based on deep reinforcement learning that is specially designed for MSTSP. Targeting both existing performance measures and our newly introduced Multi-Solution Quality Index (MSQI), our approach seeks to optimize diversity while pursuing high-quality solutions, which is featured by an encoder-decoder structured policy. Firstly, 
we design a Relativization Filter (RF) in the encoder based on both Cartesian and Polar coordinates, which improves the quality and robustness of the found solutions against variations in node distribution. 
Next, we leverage an attention-based multi-decoder architecture~\cite{xin2021multi} and further equip it with an adaptive active search mechanism, allowing the decoders to switch the baseline for balancing optimality improvement and diversity enhancement. Our method greatly improves the representation of multiple solutions and enhances the generalization capability against varying scales, while boosting the solution quality and diversity. Experimental results based on benchmark instances confirmed the superiority of our method and the effectiveness of the key designs.
\section{Related work}

\smallskip\noindent\textbf{Traditional Heuristics.}
Numerical traditional heuristics have been proposed for MSTSP. In~\cite{ronald1995finding}, GA was leveraged to yield multiple high-quality routes with a multi-chromosomal cramping design for enhanced exploration. 
Later, researchers have increasingly embraced niche techniques for addressing MSTSP, which can effectively promote diversity within a population by imposing limitations on individual similarity. This approach proves instrumental in preventing premature convergence within the solution space and facilitates algorithms to explore and maintain a wider range of diverse solutions. Consequently, many studies have integrated niche technology with other methodologies. 
For example, 1) ACO, Angus et al.~\cite{angus2006niching} applied fitness sharing and crowding to ACO to simultaneously locate and maintain multiple areas of interest in the search space; Han et al.~\cite{han2018multimodal} adopted niching strategy and multiple pheromone matrices to maintain population diversity and track the traces of multiple paths. 2) genetic algorithm, Huang et al.~\cite{huang2018seeking} combined genetic algorithm with niching technique defined in the discrete space to improve the quality and diversity of multiple solutions; 3) memetic algorithms have also been extensively studied.
The Niching Memetic Algorithm (NMA)~\cite{huang2019niching}, serves as a prominent example, utilizing parallel search for diverse and high-quality solutions. While it significantly improved computational efficiency over previous methods, its run time is still considerable. Building upon NMA, a subsequent method~\cite{do2022niching} proposed to start from an optimal solution and enhances its diversity through propagation and mutation. However, it requires the optimal solution in the first place, rendering it less flexible in many practical scenarios. \colorr{Additionally, Liu et al.~\cite{liu2021multi} extended the multi-modal single-objective TSP (i.e., the MSTSP presented in our paper) to a multi-modal multi-objective (MMTSP) context. In another recent work, Liu et al.~\cite{liu2023evolutionary} acknowledged that NMA is deemed as the state-of-the-art approach for the single-objective version of the problem, which we have adopted as benchmark in our comparative analysis.}
\begin{figure*}[t]
\centering\includegraphics[width=\linewidth]{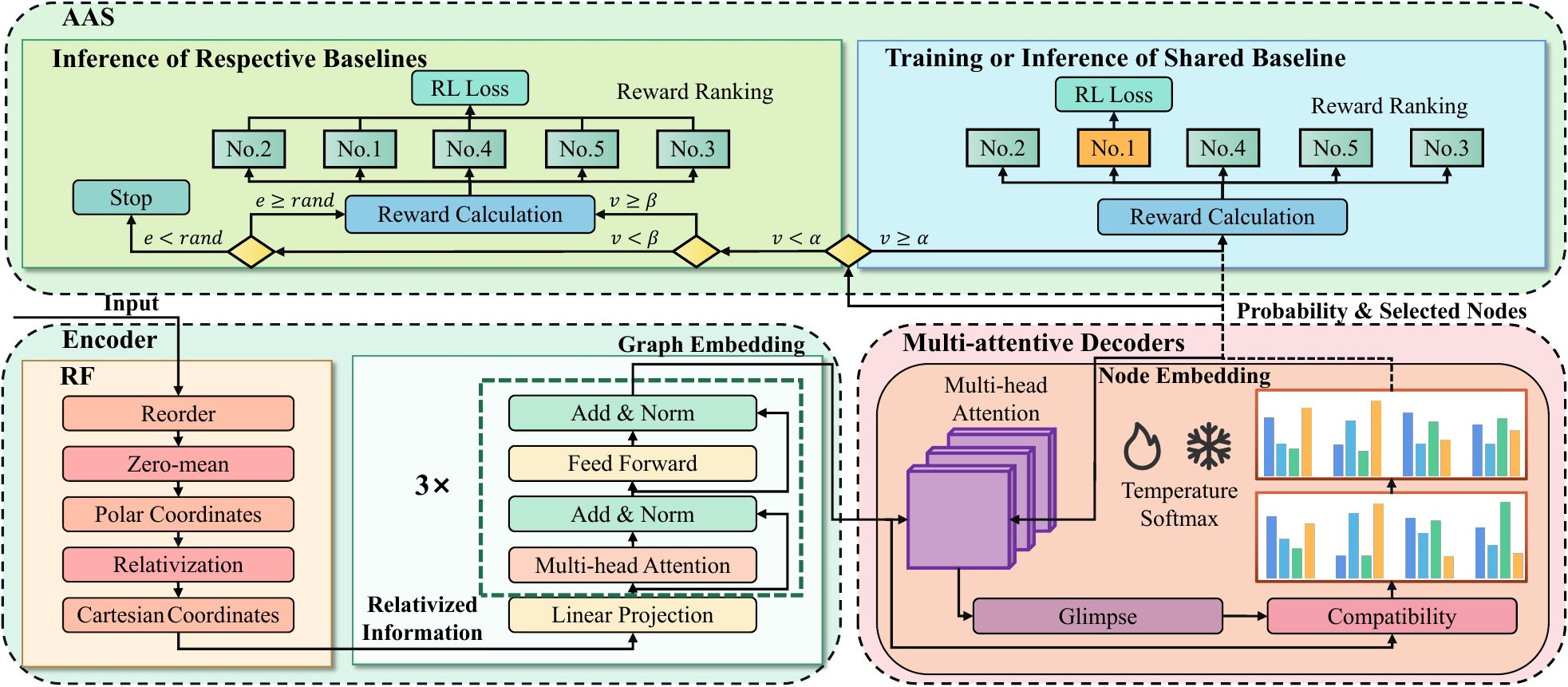}
    \caption{The architecture of our neural heuristic, RF-MA3S. It is mainly featured by a relativization filter (RF) assisted encoder, multi-attentive decoders, and the adaptive active search during inference.}
    \label{model}
\end{figure*}

\smallskip\noindent\textbf{Neural Heuristics.} Most existing neural heuristics for solving vehicle routing problems (VRPs) including the TSP focus on pursuing a single (optimal) solution. Among them, Bello et al.~\cite{bello2016neural} introduced the attention-based~\cite{bahdanau2014neural} Pointer Network (PtrNet)~\cite{vinyals2015pointer} to the actor-critic algorithm~\cite{konda1999actor} for solving TSP and 0-1 knapsack (KP). Later, the self-attention mechanism emerged and garnered high popularity in designing deep models~\cite{vaswani2017attention} for VRPs. Kool et al.~\cite{kool2018attention} adapted the work in~\cite{bello2016neural} to apply Transformer for more COPs, in which the introduced logit clipping to the decoder acts as a preset and deterministic trade-off between exploration and exploitation.
More recently, Kwon et al.~\cite{kwon2020pomo} proposed the well-known POMO, which employed two simple yet effective modifications: 
1) Multiple starting points initialization: The approach considers each point as a starting point once in parallel, and 2) Instance augmentation: The instances undergo symmetry processing, including mirroring and rotation at specific angles, to expand their quantity by eightfold.
However, the symmetry exploitation of POMO is potentially limited by the specified eight transformations only, which then motivated Kim et al.~\cite{kimsym} to further consider the problem symmetry, solution symmetry, and rotational symmetry simultaneously, improving the single solution optimality of the model. 
Many researchers have then focused on enhancing the neural heuristics based on the success of Kool et al.~\cite{kool2018attention} and Kwon et al.~\cite{kwon2020pomo}. To enhance search diversity, Xin et al.~\cite{xin2021multi} and Grinsztajn et al. ~\cite{grinsztajn2022population} improved the model from the architecture perspective, by introducing multiple decoders, with each possessing distinct parameters. To improve the optimality, several search approaches have been proposed, including sampling~\cite{bello2016neural,kool2018attention}, beam search~\cite{choo2022simulation,xin2021multi,vinyals2015pointer,joshi2019efficient}, efficient active search (EAS)~\cite{hottung2022efficient}, \colorr{flexible k-Opt~\cite{NEURIPS2023_9bae70d3}, and heavy decoder~\cite{NEURIPS2023_1c10d0c0}}. However, existing works may overlook the combined consideration of both diversity and optimization, especially for solving MSTSP.
Our method explicitly targets diverse solutions, distinguishing itself from many existing ones that mainly aim at finding a single optimal solution. Compared to active search~\cite{bello2016neural} and EAS~\cite{hottung2022efficient}, our Multi-Attentive Adaptive Active Search (MA3S) is featured by the diversity-enhancement scheme, which allows it to surpass MDAM~\cite{xin2021multi} in solution quality.

\section{MSTSP notations and measures}
\label{Performance Measures}

\subsection{MSTSP Definition}

This section will explain the definition of classical TSP, and further introduce MSTSP and the solution filters required for MSTSP. 

\noindent\textbf{\textsc{Definition 1.}}~\textbf{TSP.}
Given a complete graph $G=(V,E)$, where $V$ represents the set of nodes with size $N=|V|$, and $E$ represents the set of edges between pairs of nodes. Each edge has a weight $d(\cdot, \cdot)$ denoting its Euclidean distance.
The objective of TSP is to find the shortest Hamiltonian circle that visits each node once and returns to the first visited node. We denote a TSP solution as $\pi = [\pi_{(1)}, \cdots \pi_{(N)}]$ and the objective function as 
$\min L(\pi)=\sum_{i=1}^{N-1} d(\pi_{(i)},\pi_{(i+1)}) + d(\pi_{(N)},\pi_{(1)})$.

\noindent\textbf{\textsc{Definition 2.}}~\textbf{MSTSP.}
While sharing the same notation, MSTSP aims to pursue multiple solutions of high diversity and quality. 
For a set of solutions yielded by an algorithm, a solution filter~\cite{li2013benchmark, yu2003feature,estevez2009normalized} is typically employed to select solutions that are both high-quality and diversity. The performance indices are then calculated based on the filtered solution set. 
The solution filters in terms of quality and diversity are described below. 

\noindent\textbf{\textsc{Definition 3.}}~\textbf{Optimality Filter.} 
For any solution $\pi_i$, it must first satisfy the optimality threshold condition as $L(\pi_i)<L(\pi_{best})\cdot (1+ \delta_1)$, where $L(\cdot)$ denotes the route length, $\pi_{best}$ denotes any solution in the set with the shortest length, and $\delta_1$ represents the optimality threshold.

\noindent\textbf{\textsc{Definition 4.}}~\textbf{Diversity Filter.} Additionally, except the $\pi_{best}$, all solutions should then satisfy the diversity condition as follows, $S(\pi_i,\pi_j)=\frac{\left |\Phi\left(\pi_i\right) \cap \Phi\left(\pi_j\right)\right|}{N}< \delta_2$,
where $S$ is a similarity measure, $\Phi(\pi_i)$ denotes the set of edges for $\pi_i$, and $\delta_2$ is the similarity threshold. For the CVRP (Capacitated Vehicle Routing Problem), $N$ is replaced with $\frac{|\Phi\left(\pi_i\right)|+|\Phi\left(\pi_j\right)|}{2}$.

\subsection{Performance Measures}
For evaluating the filtered solution set of an MSTSP instance, we employ two metrics to comprehensively assess its optimality and diversity.

\noindent\textbf{\textsc{Definition 5.}}~\textbf{Diversity Indicator (DI).} 
DI~\cite{huang2018seeking} is a commonly used measure in the existing MSTSP studies. It quantifies the overlap ratio between the filtered solution set and the ground-truth optimal solution set as follows,
\begin{equation}
\operatorname{DI}(\mathbb{G}, \mathbb{S})=\frac{1}{{|\mathbb{G}|}}{\sum_{i=1}^{|\mathbb{G}|} \max_{j=1,...,|\mathbb{S}|} S\left(g_i, \pi_j\right)},
\label{eq:DI}
\end{equation}
The \( |\mathbb{S}| \) and \( |\mathbb{G}| \) represent the sizes of the solution sets \( \mathbb{S} \) and \( \mathbb{G} \) respectively, with \( \mathbb{G} \) containing the optimal solutions. The \( i^{\rm th} \) solution in \( \mathbb{G} \) is denoted by \( g_i \), and the \( j^{\rm th} \) solution in \( \mathbb{S} \) is \( \pi_j \).
Nevertheless, DI exhibits certain limitations: 1) it relies heavily on the ground-truth optimal solution set, which is however unavailable for most TSP datasets; 2) it is a unified indicator for optimality and diversity, but falls short of providing insights when there is a need to separately examine optimality and diversity achieved by the algorithms.

To complement the DI measure, we further introduce a new indicator as the second metric, i.e., Multi-Solution Quality Index (MSQI). It is developed based on Diversity Index and Optimality Index, which are described as below.
\begin{figure*}[tbh]
\centering\includegraphics[width=\linewidth]{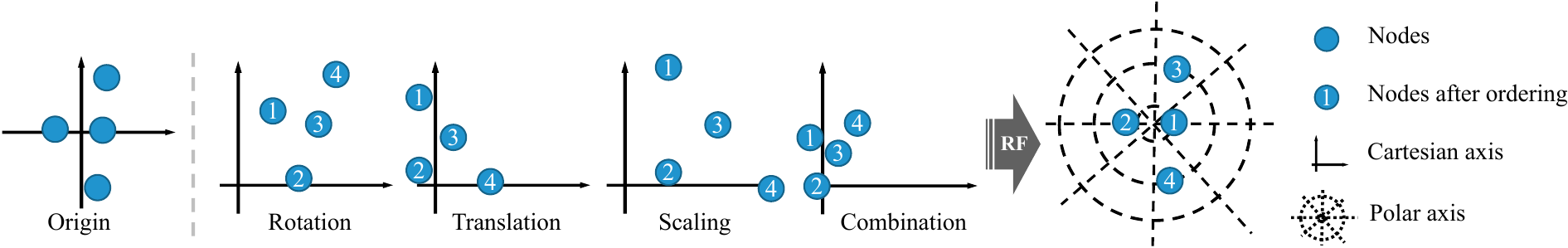}
    \caption{The multiple affine transformations of an instance become consistent after RF processing.}
    \label{RFPIC}
\end{figure*}

\noindent\textbf{\textsc{Definition 6.}}~\textbf{Diversity Index.} 
Diversity Index evaluates the \emph{diversity} between a single solution and others in the same set, which is computed as follows, 
\begin{equation} \operatorname{Diff}(\pi_i)=\frac{1}{|\mathbb{S}|-1} \sum_{j=1}^{|\mathbb{S}|} U\left(\pi_i, \pi_j\right). \end{equation}
Here, the function $U$ is computed using the similarity $S$ as follows,
\begin{equation}
U(\pi_i, \pi_j)=
\begin{cases}
2(1- {S(\pi_i, \pi_j)}) , \quad  &\frac{1}{2}< S(\pi_i, \pi_j)\leq 1 , \\
1 , \quad &0\leq S(\pi_i, \pi_j)\leq \frac{1}{2},
\end{cases}\label{fu}
\end{equation}
where we let $U(\pi_i, \pi_j)=1$ if more than half of the edges of solutions $\pi_i, \pi_j$ differ, suggesting a high degree of difference; Otherwise, we assign a smaller value $0\leq U(\pi_i, \pi_j)<1$ according to $S(\pi_i, \pi_j)$. 
\noindent\textbf{\textsc{Definition 7.}}~\textbf{Optimality Index.} 
The \emph{optimality} of a single solution is then measured based on the normalized distance within a threshold between the route length of this solution and the route length of the optimal solution as follows, 
\begin{equation}\operatorname{Opt}(\pi_i) = \frac{(1+\delta_1)\cdot L({\pi_{best}})-L({\pi_i})}{\delta_1 \cdot L({\pi_{best}})}.
\label{eq:best}
\end{equation}

\noindent\textbf{\textsc{Definition 8.}}~\textbf{Multi-Solution Quality Index (MSQI).}
Before reaching MSQI, we first present the SQI of a single solution, which is computed according to Eq.~(\ref{eq:MSQI1}). 
In general, the harmonic mean is sensitive to extreme values. Therefore, when there is only one solution in set $\mathbb{S}$, the difference is 0, and the SQI will also become 0. Similarly, poor optimality of the solution will also lead to a low SQI. 
MSQI aggregates the SQI of each solution $\pi_i \in \mathbb{S}$. It explicitly measures both optimality and diversity, without requiring a ground-truth optimal solution set. A higher MSQI or SQI value signifies a solution set or a solution that excels in both optimality and diversity.
\begin{equation}
\operatorname{SQI}(\pi_i)=\frac{2}{ \frac{1}{\text{Opt}(\pi_i)}+\frac{1}{\text{Diff}(\pi_i)}}, 
\operatorname{MSQI}=\frac{|\mathbb{S}|}{\sum_{i}^{|\mathbb{S}|}\frac{1}{\text{ SQI}(\pi_i)}}. 
\label{eq:MSQI1}
\end{equation}

\section{Methodology}

We introduce Relativization Filter assisted Multi-Attentive Adaptive Active Search (RF-MA3S) for MSTSP. As depicted in Figure \ref{model}, its architecture follows the encoder-decoder structure~\cite{kwon2020pomo}, featuring an RF-assisted encoder and multi-attentive decoders. After training via reinforcement learning to learn transferable features, we use an adaptive active search strategy for each test instance during inference. While the model learned via reinforcement learning explicitly outputs multiple solutions, the per-instance active search aims to balance the optimality and diversity of MSTSP more effectively.

\subsection{Relativization Filter (RF) Assisted Encoder}

For many COPs including the TSP, an instance and its affine transformations such as translation, rotation, scaling, mirroring, etc, often share the same optimal solution. While prior attempts have sought to capture such invariance through data augmentation~\cite{kwon2020pomo} and auxiliary losses~\cite{kimsym}, we offer a simpler yet effective alternative by explicitly embedding the invariance using a unified representation. 
This may bolster both solution optimality and diversity in MSTSP, as it minimizes redundant solution representations throughout both the training and inference processes.
Notably, it improves upon previous efforts by reducing the reliance on instance augmentation during inference~\cite {kwon2020pomo,kimsym}.

To achieve this, we propose exploiting Relativization Filter (RF) in the encoder, which adopts a series of relativization operations to convert the absolute information of the nodes into relative information while capturing the internal relationships among the nodes. As illustrated in Figure \ref{RFPIC}, this filter helps screen out the instances which are essentially the affine transformations of others since they share the same eventual representation, and thus enhance the robustness of the encoder. Given a set of nodes with Cartesian coordinates $(x_1,y_1),\dots,(x_N,y_N)$ as the input, we execute our RF as follows.

\smallskip\noindent\textbf{Reorder.} To alleviate the order influence of the input nodes, they are sorted first by their $y$ values, then by $x$ values in descending order within the Cartesian coordinates. The relative positions of nodes with the same values will not be changed after sorting, resulting in a unique sequence of nodes.

\smallskip\noindent\textbf{Zero-mean.} All $x$ and $y$ subtract the respective means, and lead to $x'$ and $y'$. It moves the centroid of the instance to the origin, and helps capture the influence of translation.

\smallskip\noindent\textbf{Conversion to Polar Coordinates.} Polar coordinates $(\rho,\theta)$ are derived via $\arctan(\frac{y'}{x'})$ and $\sqrt{x'^2+y'^2}$. However, such $\theta$ cannot preserve the quadrant information of the Cartesian coordinate system, as its range is limited to $[-\frac{\pi}{2},\frac{\pi}{2}]$. We thus add $+\pi$ to $\theta$ for nodes located in the second and third quadrants of the original Cartesian coordinate system, expanding the range to $[-\frac{\pi}{2},\frac{3\pi}{2}]$.

\smallskip\noindent\textbf{Polar Coordinates Relativization.} First, it normalizes $\rho$ as $\rho'=\frac{\rho}{\rho_{\max}}$, which helps capture the scaling effect. Then, it sorts the polar coordinates $(\rho',\theta)$ in the descending order of $\rho'$, and normalizes angles by subtracting the first angle, mitigating the rotational effect.

\smallskip\noindent\textbf{Conversion to Cartesian Coordinates.} The computed values of $(\rho', \theta')$ are converted back into Cartesian coordinates $(x'', y'')$ by $x''=\rho'\cos{\theta'}$ and $y''=\rho'\sin{\theta'}$, 
and fed to the encoder.

The above procedure might be less effective in handling the mirroring transformation. Therefore, the $\times 2$ instance augmentation \cite{kwon2020pomo} is applied during inference, where an instance with input $(x,y)$ and its mirroring instance with input $(y,x)$ are considered. Note that other types of mirroring can be regarded as a combination of this mirroring and the aforementioned translation and rotation.

\subsection{Multi-attentive Decoders}

Inspired by the architecture in the Multi-Decoder Attention Model (MDAM)~\cite{xin2021multi}, we reduce the number of encoder blocks to three and expand the attention-based single decoder in POMO~\cite{kwon2020pomo} with five duplicates. 
In MDAM~\cite{xin2021multi}, solution diversity is bolstered through the employment of KL divergence, which, however, incurs heavy computation overhead. In our approach, we simply deploy the five decoders of the same structure in parallel. Note that, on the one hand, due to the mechanism of multiple starting points, even a single decoder is able to produce multiple solutions. On the other hand, the random initialization of the parameters for each decoder will also boost the diversity of the solutions. Furthermore, most importantly, the subsequent adaptive active search will further diversify those decoders by adaptively updating their parameters while pursuing high-quality solutions.
\subsection{Training Phase}
Our training algorithm mainly follows that of POMO~\cite{kwon2020pomo}. Specifically, for a batch of input instances $s$, each decoder independently performs parallel computations with multiple starting points to obtain the REINFORCE loss with the greedy rollout as follows,
\begin{equation}\resizebox{.9\hsize}{!}{$\nabla_\theta J(\theta) = \frac{1}{BDN}\sum_{m}^{B}\sum_{j}^{D}\sum_{i}^{N} (L(\pi_{i,j}^m |s)-b(s))\nabla_\theta \log p_\theta (\pi_{i,j}^m |s)$.}
\end{equation}
Here, the result of the best decoder (with shortest averaged length) in current epoch is taken as the baseline $b(s)$. Namely, 
$b(s)=\frac{1}{N} \sum_{i}^{N} (L(\pi_{i,l}|s))$, where $l = \mathop{\arg\min}\limits_{j}(L(\pi_{\cdot ,1}|s),
\cdots, L(\pi_{\cdot ,D}|s))$.

\smallskip\noindent\textbf{Temperature Softmax.} During the training phase, we also employ the temperature softmax~\cite{yi2019sampling} as a replacement for logit clipping in the decoder~\cite{bello2016neural} expressed as 
$ {\rm Softmax}\left(a_{i,j},\tau\right)=\frac{e^{\frac{a_{i,j}}{\tau}}}{\sum_{j'} e^{\frac{a_{i,j'}}{\tau}}}$. 
Here, $\tau=\frac{\tau_0}{1+\log _{10} T}$, $T$ denotes the current epoch, and $\tau_0$ is a constant and set to 2. The temperature softmax is used to enhance model exploration by setting a high initial temperature, reducing the disparity in probability values for selecting the next node. As training progresses, temperature drops, enhancing probability differences and aiding convergence.

\subsection{Adaptive Active Search Phase}

After the training phase, we utilize our proposed Adaptive Active Search (AAS) during inference to synergize with the multi-attentive decoders, so as to pursue diverse yet high-quality solutions. In each iteration, it first samples the solutions for a given test instance using the respective decoders and calculates the corresponding loss w.r.t a baseline. Then, the parameters of the model are adjusted based on the loss, aiming to increase the likelihood of yielding high-quality solutions in the subsequent iterations. In comparison with AS~\cite{bello2016neural} and EAS~\cite{hottung2022efficient}, our AAS is able to adaptively determine the switching of the baseline for parameter update according to the convergence status of the model, and ensure the diversity of multiple solutions. We consider adjusting all parameters of the pre-trained models for better performance. Moreover, a designed termination condition for the iterations will also potentially prevent the diversity deterioration caused by excessive iterations. 
\begin{algorithm}[t]

\small
		\begin{algorithmic}[1]
			\Require~Instance $s$, trainable parameter $\theta$, batch size $B$, number of decoders $D$, number of starting points $N$, baseline switching threshold $\alpha$, probabilistic stopping threshold $\beta$, maximum iterations $T$;			
			\Ensure A solution set $\mathbb{S}$.
			\State Initialize $ L \leftarrow +\infty; \mathbb{S} \leftarrow \emptyset; switch\leftarrow 0$;	  
			\For{$t=1, 2, \cdots , T$} 
			\State \resizebox{.87\hsize}{!}{$\pi_{i,j}^m \sim \operatorname{SAMPLE}(p_{\theta}(\cdot | s))$ for $m$, $j$, $i$ $\in$ $\{1,...,B\}$, $\{1,...,D\}$, $\{1,...,N\}$};  
			\State $l \leftarrow  \mathop{\arg\min}(L(\pi_{\cdot ,1}|s),...,L(\pi_{\cdot ,D}|s))$;	   
                \State $temp \leftarrow  \frac{1}{BDN}\sum_{m}^{B}\sum_{j}^{D}\sum_{i}^{N}(L_{i,j}^m(\pi|s))$;	
                \State \textbf{If} $L > temp$, $L \leftarrow temp $ and $ \mathbb{S} \leftarrow \pi$; 
                \State \textbf{If} {$switch$}, $b(s)\leftarrow \frac{1}{N} \sum_{i}^{N} (L(\pi_{i,\cdot}|s))$; 
                \State \textbf{else}, $b(s)\leftarrow\frac{1}{N} \sum_{i}^{N} (L(\pi_{i,l}|s))$;
   
			\State \resizebox{.87\hsize}{!}{$\nabla_\theta J(\theta) \leftarrow \frac{1}{BDN}\sum_{t}^{B}\sum_{j}^{D}\sum_{i}^{N} (L(\pi_{i,j}^m  |s)-b(s))\nabla_\theta \log p_\theta (\pi_{i,j}^m |s)$};   
			\State $\theta \leftarrow $ ADAM$(\theta,g_\theta)$;  
                \State $f \leftarrow \frac{\nabla_\theta J(\theta)}{J(\theta)}$; 
                \State $e \leftarrow \frac{{\beta}-f}{{\beta}}$;   

                \State \textbf{If} $f <\alpha$, $switch \leftarrow 1$; meanwhile, \textbf{If} $e<rand$, break.
			\EndFor	
                    
			\State \Return $\mathbb{S}$
    \end{algorithmic}
    \caption{Adaptive Active Search}
    \label{alg2}
\end{algorithm}

\smallskip\noindent\textbf{Adaptive Baseline.} Two types of baselines are involved in the active search phase and described as follows. 1) The results of the best decoder (with the shortest averaged route length) are used as the \emph{shared} baseline. It will encourage the convergence of all decoders towards the global optimum, enhancing the optimality with a relatively large total loss; 2) The results of each individual decoder (the averaged route length) are used as their \emph{respective} baselines. It will encourage the convergence of each decoder towards a potential local optimum, enhancing the diversity with a relatively small total loss. Particularly, at the early stage, we need to emphasize on the quality of solutions and ensure the movement towards the optimum, where the \emph{shared} baseline is preferred. Afterwards, the diversity of the solutions should be guaranteed and strengthened, where the \emph{respective} baseline is more desired. In our AAS, this switching decision is adaptively made based on the convergence degree of the model,
\begin{equation}
 f= \frac{\bigtriangledown_\theta J(\theta)}{J(\theta)} , 
\end{equation}
where $J(\theta)$ is the objective function (the negative value of the route length), divided for normalization w.r.t the objective distribution scale. The switch will be incurred if $f<\alpha$, where $\alpha$ is a positive constant as the threshold.

\begin{table*}[t]\setlength{\tabcolsep}{6pt}
\caption{Experiment results on MSTSPLIB.}
\resizebox{0.9\textwidth}{!}{%
\begin{tabular}{@{}l|cc|cc|cc|cc|ccc@{}}
\toprule
\hline
\multirow{3}{*}{Method} & \multicolumn{2}{c|}{1st category} & \multicolumn{2}{c|}{2nd category} & \multicolumn{2}{c|}{3rd category} & \multicolumn{2}{c|}{4th category} & \multicolumn{3}{c}{Entire test set}  \\ 
 & \multicolumn{2}{c|}{(9 - 12)} & \multicolumn{2}{c|}{(10 - 15)} & \multicolumn{2}{c|}{(28 - 33)} & \multicolumn{2}{c|}{(35 - 66)} & \multicolumn{3}{c}{(9 - 66)}  \\
 & MSQI $\uparrow$ & DI $\uparrow$ & MSQI $\uparrow$ & DI $\uparrow$ & MSQI $\uparrow$ & DI $\uparrow$ & MSQI $\uparrow$ & DI $\uparrow$ & \begin{tabular}[c]{@{}c@{}}MSQI\\ Gap(\%) $\downarrow$\end{tabular} & \begin{tabular}[c]{@{}c@{}}DI\\ Gap(\%) $\downarrow$\end{tabular} & Time $\downarrow$ \\ \midrule
NGA & 0.837 & 0.932 & 0.907 & 0.909 & 0.709 & 0.883 & 0.783 & 0.748 & 1.791 & 7.984 & 13.5H  \\
NMA & 0.856 & 1.000 & 0.932 & 0.953 & 0.698 & 0.942 & 0.801 & 0.853 & 0.000 & 0.000 & 40.0M  \\ 
\hline
POMO (20) & 0.789 & 0.946 & 0.707 & 0.820 & 0.863 & 0.714 & 0.722 & 0.655 & 8.649 & 16.455 & 8.0S  \\
Sym-NCO (20) & 0.804 & \textbf{0.967} & 0.672 & 0.843 & 0.845 & 0.739 & 0.693 & 0.657 & 10.870 & 14.805 & 8.0S  \\
MDAM-greedy (20) & 0.356 & 0.756 & 0.544 & 0.711 & 0.791 & 0.545 & 0.731 & 0.516 & 26.917 & 32.538 & 3.0M  \\
MDAM-bs (20) & 0.596 & 0.886 & 0.437 & \textbf{0.876} & 0.850 & 0.772 & \textbf{0.798} & 0.724 & 19.027 & 12.895 & 15.0M  \\
EAS (20) & 0.606 & 0.962 & 0.619 & 0.844 & 0.642 & 0.763 & 0.456 & 0.669 & 32.327 & 14.060 & 42.0M  \\
RF-MA3S (20) & \textbf{0.805} & 0.953 &\textbf{0.708} & 0.836 & \textbf{0.877} & \textbf{0.862} & 0.791 & \textbf{0.832} & \textbf{5.157} & \textbf{6.406} & 30.0M  \\ \hline
POMO (50) & {0.617} & 0.975 & 0.679 & 0.812 & 0.825 & 0.742 & 0.715 & 0.674 & 15.503 & 14.691 & 8.0S  \\
Sym-NCO (50) & 0.553 & 0.970 & 0.779 & 0.844 & 0.756 & 0.739 & 0.649 & 0.655 & 18.645 & 14.812 & 8.0S  \\
MDAM-greedy (50) & 0.434 & 0.590 & 0.231 & 0.477 & 0.667 & 0.532 & 0.321 & 0.416 & 53.919 & 46.986 & 3.0M  \\
MDAM-bs (50) & 0.460 & 0.911 & 0.486 & 0.832 & 0.624 & 0.678 & 0.493 & 0.666 & 39.157 & 17.247 & 16.0M  \\
EAS (50) & 0.536 & \textbf{0.980} & 0.550 & 0.825 & 0.534 & 0.708 & 0.287 & 0.680 & 45.785 & 14.572 & 48.0M  \\
RF-MA3S (50) & \textbf{0.657} & 0.976 & \textbf{0.791} & \textbf{0.914} & \textbf{0.840} & \textbf{0.815} & \textbf{0.808} & \textbf{0.793} & \textbf{6.794} & \textbf{6.164} & 33.0M  \\ \hline
LEHD (100)& \textbf{0.683} & 0.859 & 0.504 & 0.661 & 0.490 & 0.706 & 0.497 & 0.734 & 34.616 & 19.909 & 1.4M \\
POMO (100) & 0.305 & 0.934 & 0.438 & 0.795 & 0.659 & 0.729 & 0.541 & 0.658 & 42.282 & 17.058 & 8.0S  \\
Sym-NCO (100) & {0.472} & 0.881 & \textbf{0.531} & 0.701 & 0.742 & 0.715 & 0.489 & 0.667 & 35.359 & 20.747 & 8.0S  \\
MDAM-greedy (100) & 0.000 & 0.085 & 0.000 & 0.145 & 0.527 & 0.317 & 0.252 & 0.426 & 78.909 & 72.024 & 3.0M \\
MDAM-bs (100) & 0.290 & 0.762 & 0.381 & 0.770 & 0.545 & 0.454 & 0.167 & 0.284 & 62.788 & 41.407 & 15.0M  \\
EAS (100) & 0.437 & 0.936 & 0.509 & 0.800 & 0.492 & 0.713 & 0.228 & 0.585 & 53.236 & 20.011 & 1.5H \\
RF-MA3S (100) & 0.385 & \textbf{0.992} & 0.502 & \textbf{0.877} & \textbf{0.858} & \textbf{0.766} & \textbf{0.785} & \textbf{0.802} & \textbf{23.683} & \textbf{7.161} & 1.0H  \\ \midrule\bottomrule
\end{tabular}%
}\label{SSP}
\end{table*}
\smallskip\noindent\textbf{Adaptive Termination.} Moreover, rather than using a simple threshold to terminate the whole iterations, we use a probability to represent the termination condition, so as to counteract the random factors in each iteration. In particular, let $e$ denote the probability of early termination, and let $\beta = 0.5\alpha$, we compute such probability in Eq.~(\ref{eq:stop}) and summarize the procedure of the proposed adaptive active search in Algorithm~\ref{alg2}.
\begin{equation}
e=\left\{\begin{array}{cc}
\frac{\beta-f}{\beta}, & f<\beta ,\\
0, & f \geq \beta.
\end{array}\right.
\label{eq:stop}
\end{equation}

\colorr{The absolute value of the gradient consistently decreases as iterations progress, and the probability of iteration termination converges to 1 once it falls below the predetermined threshold.}
\section{Experiments}

In our experiments, we evaluated the multi-solution performance of RF-MA3S in both TSP and CVRP, comparing it with existing traditional heuristic and neural heuristic methods.
\subsection{Experiment Settings}

\smallskip\noindent\textbf{Datasets.} To comprehensively evaluate the proposed method, we conduct experiments on MSTSPLIB~\cite{huang2018seeking}, TSPLIB~\cite{reinelt1991tsplib}, CVRPLIB~\cite{CVRPLIBdata}, and the uniformly distributed synthetic instances as used in~\cite{kwon2020pomo, kimsym}.
Regarding the instances in MSTSPLIB, they are labeled as mstsp1 – mstsp25, and categorized into four groups based on their distributions.
Regarding the TSPLIB and CVRPLIB, they include widely used practical instances for TSP and CVRP, respectively. 
Regarding the uniform instance set, it was only used in the experiment for investigating the affine transformation resistance effects.

\smallskip\noindent\textbf{Competitors.} 
The proposed method is named RF-MA3S. Further, we compare it with 
a few traditional heuristic algorithms and neural heuristic baselines. 
The heuristic algorithms include NGA~\cite{huang2018seeking} (only tested on MSTSP becasuse of its excessive time demands) and the state-of-the-art NMA~\cite{huang2019niching}. The neural baselines involve POMO~\cite{kwon2020pomo}, Sym-NCO~\cite{kimsym}, EAS (with pre-trained model)~\cite{hottung2022efficient}, MDAM with greedy rollout (MDAM-greedy), 50-width beam search (MDAM-bs)~\cite{xin2021multi}, and \colorr{LEHD with RRC (trained on instances with a default scale
of $N = 100$)}~\cite{NEURIPS2023_1c10d0c0}. All experiments are conducted on a machine with NVIDIA RTX 3090 GPU and Intel Core i9-10980XE CPU. 
Code is available: \href{https://github.com/LiQisResearch/KDD--RF-MA3S}{https://github.com/LiQisResearch/KDD--RF-MA3S}.
\subsection{Hyperparameter} 
NGA~\cite{huang2018seeking} and NMA~\cite{huang2019niching} are tested for 10 independent runs. 
The neural methods are trained on uniformly distributed datasets for 100 epochs, each with 10k instances, using a batch size of 64.

For the active search during inference, our MA3S adopts the switching threshold as $\alpha=0.005$ and a learning rate of $1\times 10^{-5}$. To guarantee the inference speed, we set a maximum number of iterations to 2000 when testing on TSPLIB~\cite{reinelt1991tsplib} and CVRPLIB~\cite{CVRPLIBdata}. For a fair comparison, the iteration count for EAS is set to five times the actual iteration count of MA3S, in light of the five decoders in our approach. For the LEHD~\cite{NEURIPS2023_1c10d0c0}, relying solely on greedy rollout does not yield multiple solutions; hence, the LEHD employed in this paper incorporates the RRC~\cite{NEURIPS2023_1c10d0c0}technology (a versatile technique applicable to various neural methods, with setting of 50). Unless otherwise stated, the settings of the compared methods follow the recommendations in their original papers. For solution filtering, considering the varied difficulties in seeking the optimality and diversity on different test sets, the optimality threshold $\delta_1$ and similarity threshold $\delta_2$ are adjustable, while ensuring that all methods use the same settings on the same test set. Empirically, we use $\delta_1=0.1$ and $\delta_2=0.8$ for MSTSPLIB~\cite{huang2018seeking}, $\delta_1=0.1$ and $\delta_2=0.9$ for TSPLIB, $\delta_1=0.2$ and $\delta_2=0.8$ for CVRPLIB and real-world dataset. 

\begin{table*}[t]
\setlength{\tabcolsep}{8pt}
\centering
\caption{Experiment results on TSPLIB.}
\resizebox{.9\textwidth}{!}
{%
\begin{tabular}{@{}l|cccccccc|c@{}}
\toprule\midrule
{Method} & eil51 & berlin52 & st70 & pr76 & kroA100 & lin105  & rd400 & rat783  & Time (eil51-lin105) $\downarrow$  \\ \midrule
NMA & 0.551 & 0.466 & 0.586 & 0.480 & 0.366 & 0.496 & 0 & 0  & 1.7H   \\ \hline
LEHD &  0.273 & 0.000  &  0.000 & 0.000  &  0.000 & 0.000 & 0.281 & 0.334  & 14.0M  \\ 
POMO & 0.632 & 0.100 & 0.630 & 0.515 & 0.158 & 0.156 & 0 & 0  & 3.0S \\
Sym-NCO & 0.630 & 0.189 & 0.642 & 0.529 & 0.134 & 0.210  & 0 & 0  & 3.0S \\
MDAM-greedy & 0.546 & 0.000 & 0.560 & 0.336 & 0.073 & 0.000  & 0 & 0 & 0.91M \\
MDAM-bs & 0.224 & 0.110 & 0.477 & 0.440 & 0.410 & 0.289 & 0.364 & 0   & 11.0M \\
EAS & 0.519 & 0.010 & 0.625 & 0.410 & 0.183 & 0.142 & 0.070 & 0  & 15.0M  \\
RF-MA3S & \textbf{0.637} & \textbf{0.428} & \textbf{0.653} & \textbf{0.577} & \textbf{0.505} & \textbf{0.535} & $\mathbf{0.578}$ & $\mathbf{0.567}$ & 6.6M \\ \midrule\bottomrule
\end{tabular}%
}\label{TSPLIB}
\end{table*} 

\subsection{Performance Comparisons}
\smallskip\noindent\textbf{Results on MSTSPLIB.}
The results are shown in Table~\ref{SSP}, where the MSQI, DI, and inference time of different algorithms on the four categories of MSTSPLIB are reported. The last column reports the overall results, where the performance gaps are computed w.r.t. NMA. The best results in each group of comparison are marked in bold. It is worth noting that the neural models are trained once and then directly applied to solve instances with varying node scales. For example, the result of POMO (20) means applying the POMO model trained on instances of size 20 to infer the 1st category of MSTSP instances with sizes ranging from 9 to 12. It can be observed that RF-MA3S outperforms other neural methods significantly across various categories of instances by using different training scales. Generally, RF-MA3S (20) performs better than RF-MA3S (50) and RF-MA3S (100) since the node scale (20) used to train this model is close to the scales of many instances in MSTSPLIB. Then, the results of RF-MA3S (20) on the entire set show an MSQI gap of 5.157\% and a DI gap of 6.406\% compared to the state-of-the-art NMA, while the testing time is only about three fourths of NMA. Note that a closer match between the instance scales of training and testing sets is more favourable for improving model efficiency. But it can be observed that RF-MA3S is less affected by these scale differences when compared to other neural heuristic methods, demonstrating its greater robustness. For the other methods, Sym-NCO and POMO perform similarly. EAS and LEHD exhibit good DI values but much lower MSQI values. By looking into the details, we found that it is the relatively inferior diversity-seeking capability of EAS that reduces the MSQI. MDAM yields a limited number of solutions derived from greedy inference, while the solutions obtained through beam search have high similarity, resulting in relatively low diversity in solutions.

\smallskip\noindent\textbf{Results on TSPLIB.}
\colorr{We have also evaluated our model on a set of instances from the TSPLIB, including eil51, berlin52, st70, pr76, kroA100, lin105, rd400, and rat783. All models used in this experiment were trained on instances with a default scale of $N=100$. Because the ground-truth solution set $\mathbb{G}$ in Eq. (\ref{eq:DI}) is unavailable, it is challenged to calculate the DI measure for TSPLIB. We hence focus on the MSQI and inference time. In Table~\ref{TSPLIB}, the performance of NMA decreases in comparison with its results on MSTSPLIB, due to the inferior scalability of traditional diversity optimizers. On the TSPLIB, RF-MA3S performs even better than NMA. For the inference time, NMA consumes 1.7H (hour), whereas the RF-MA3S consumes 6.6M (minute). The efficiency of RF-MA3S is about 15 times higher than that of NMA, owing to its desirable performance, especially the proposed adaptive termination strategy in MA3S. While most methods, including the LEHD for large-scale single-solution problems, show inadequate MSQI on instances rd400 and rat783, RF-MA3S maintains superior performance.}

\begin{table}[tb]
\setlength{\tabcolsep}{0.8pt}
\centering
\caption{Affine resistance performance (\%).}
\resizebox{.475\textwidth}{!}{%
\begin{tabular}{@{}l|ccc|ccc@{}}
\toprule\midrule
\multirow{2}{*}{Affine}  & \multirow{2}{*}{POMO} & \multirow{2}{*}{Sym-NCO} & \multirow{2}{*}{POMO-RF} & POMO &Sym-NCO &  POMO-RF  \\ 
& & & &  ×8&×8 &×2 \\
\midrule
Translation  & 0.105 & 0.112 & \textbf{0}  & 0.009 & 0.014 & \textbf{0}   \\
Rotation  & 0.060 & 0.014 & \textbf{0}& 0.011 & -0.002 & \textbf{0}   \\
Scaling  & 29.998 & 83.399 & \textbf{0}  & 16.450 & 58.744 & \textbf{0}   \\
Mirroring  & -0.002 & 0.004 & -0.002 & \textbf{0} & \textbf{0} & \textbf{0}  \\
Mixture  & 36.122 & 82.185 & -0.002  & 21.798 & 57.250 & \textbf{0}   \\ \midrule\bottomrule
\end{tabular}%
}\label{tablerf}
\end{table}

\subsection{In-depth Analysis of RF-MA3S}

The following models are trained on instances with a default scale of $N=50$.

\smallskip\noindent\textbf{Affine Transformation Resistance Performance by RF.} Five types of affine transformation experiments are conducted, where the test instances are undergone translation, rotation, scaling, mirroring, and a combination of the above four types, respectively. For translation, the coordinates of nodes in an instance are shifted by a random value between $[-10,10]$. For rotation, the nodes are rotated randomly around the centre position of all nodes. For scaling, the distance between nodes is enlarged by 100 times and not normalized (in this case we divide the cost by 100 when calculating the Gap). For mirroring, the $x$ and $y$ coordinates of the same group of points are swapped. For the mixture of transformations, all four types of affine transformations are applied simultaneously.

In the subsequent analysis, POMO-RF refers to POMO method integrated with our RF, and Sym-NCO enforces symmetry by minimizing the symmetric loss function. We evaluated these configurations, both with and without instance augmentation. 
Table~\ref{tablerf} exhibits the affine transformation resistance performance of different methods, by showing the gap of results on transformed instances compared to original results. The Gap of ``0'' are highlighted in bold. 
Without instance augmentation, the POMO-RF method demonstrates superior stability compared to POMO and can resist the effects of translation, rotation, and scaling perfectly, with minimal impact from the mirror transformation. In contrast, both POMO and Sym-NCO methods are affected by each type of affine transformation, resulting in certain gaps (positive or negative). Scaling transformation has a significant impact on both methods (especially Sym-NCO), with gaps reaching up to 29.998\% and 83.399\%, respectively. With $\times 2$ instance augmentation, the RF method further resists the mirror transformation. The absolute values of gaps of POMO and Sym-NCO are relatively smaller than that without augmentation, but still far behind the proposed POMO-RF.

\begin{figure}[tbh] 
    \centering
    \includegraphics[width=.9\linewidth]{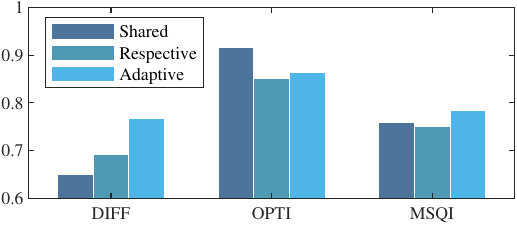}
    \caption{Investigating adaptive design in MA3S.}
    \label{T4toF}
\end{figure}
\smallskip\noindent\textbf{Optimality and Diversity Trade-off in MA3S \colorr{(AAS vs AS)}.}
As mentioned in Section~\ref{Performance Measures}, MSQI is composed of two sub-indices that evaluate the diversity and optimality, respectively. While our MA3S employs the adaptive baseline used in different decoders, here we investigate the baseline adaptation mechanism with respect to diversity and optimality performance. 
First, we removed AAS but integrated AS to create a new model for comparison, which we name RF-MDAS. Then, for better comparison between RF-MA3S and RF-MDAS, we consider two distinct types of baselines for the latter: the \emph{shared} baseline, which is based on the results of the best decoder, and the \emph{respective} baselines, which are based on the results of each individual decoder.
Figure~\ref{T4toF} shows results from a consistent iteration limit of $5 \times \text{Problem Size}$. It indicates that RF-MDAS with a shared baseline excels in optimality, while the respective baseline version improves in diversity. Meanwhile, the \emph{adaptive} baseline balances diversity and optimality, achieving the highest MSQI.

\begin{figure}[tbh] 
    \centering
    \includegraphics[width=\linewidth]{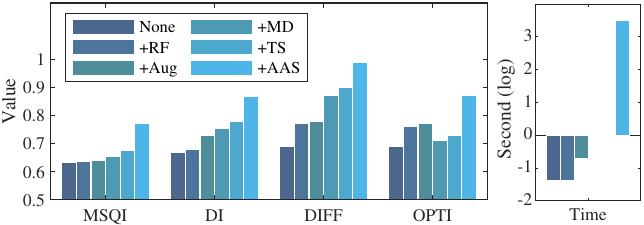}
    \caption{Ablation studies. Note: each experiment is incrementally adding components based on the previous one.}
    \label{T5toF}
\end{figure}

\smallskip\noindent\textbf{Ablation Study.}
We further evaluate the contributions of our major components by superposing the RF, $\times 2$ instance augmentation (Aug), multi-decoder (MD), temperature softmax (TS), and AAS in a progressive manner, as shown in Figure~\ref{T5toF}. The RF technique reduces the MSQI value but improves DI, thus increasing the overlap between the solution set and the set of ground-truth optimal solutions. The $\times 2$ instance augmentation doubles the inference time, but it significantly enhances performance. The multi-decoder method, utilizing five parallel decoders for problem-solving, brings a substantial increase in the number of obtained solutions, leading to steady growth in both DI and MSQI. Additionally, the use of temperature softmax helps balance the exploration and exploitation of the model, which further improves both measures. AAS has dual effects: it improves solution quality, increases the number of solutions in the filtered set, and enhances the model's exploration-exploitation trade-off, resulting in significant improvements in MSQI and DI. However, the AAS also has the side effect of longer inference time.

\begin{table}[htb]\setlength{\tabcolsep}{15pt}
\caption{The effect of threshold $\alpha$ on MSQI performance.}
\centering
\resizebox{0.475\textwidth}{!}{%
\begin{tabular}{@{}l|c|cc}
\toprule \midrule
{Method}                & {$\alpha$} & {MSQI $\uparrow$ } & {Time$\downarrow$} \\ \midrule
NMA                            & -          & 0.491          & 1.7H           \\
EAS (100)                      & -          & 0.314          & 15.0M            \\ \hline
\multirow{5}{*}{RF-MA3S (100)} & 0.005      & 0.560          & 6.6M           \\
                               & 0.0075     & 0.523          & 5.1M           \\
                               & 0.01       & 0.508          & 3.8M           \\
                               & 0.02       & 0.393          & 2.3M           \\
                               & 0.05       & 0.267          & 13.0S          \\ \midrule\bottomrule
\end{tabular}}\label{alphachange2}
\end{table}

\smallskip\noindent\textbf{Impact of the $\alpha$ in AAS.}
To more intuitively demonstrate the trade-off between efficiency and accuracy in our method, we conducted additional experiments on TSPLIB by adjusting the $\alpha$ value within the range of \{0.005, 0.0075, 0.01, 0.02, 0.05\}. Table~\ref{alphachange2} shows that an increase in \(\alpha\) decreases inference time by prompting earlier termination of iterations in the AAS. excessively premature exits may compromise MSQI performance. At \(\alpha = 0.01\), the RF-MA3S (100) model achieves better MSQI than the baseline NMA and requires only 3.8M for inference, making it about 27 times more efficient.

\begin{table}[t]
\setlength{\tabcolsep}{0.8pt}
\centering
\caption{Experiment results on CVRPLIB.}
\resizebox{.475\textwidth}{!}{%
\begin{tabular}{@{}l|c|ccccccc@{}}
\toprule\midrule
\multirow{2}{*}{Dataset}  & \multirow{2}{*}{Cost} & \multirow{2}{*}{POMO} & {Sym} & \multirow{2}{*}{LEHD} & MDAM & MDAM & \multirow{2}{*}{EAS} &{RF} \\ 
&  &  & -NCO&  & -greedy & -bs & &-MA3S \\ 
\midrule
A-n32-k5   & 784 & 0.260   & 0.342 & \textbf{0.685} & 0.000 & {0.588}& 0.229                     & 0.331 \\
A-n33-k5   & 661 & 0.518   & 0.142 & \textbf{0.897} & 0.597 & 0.432  & 0.606                     & {0.726} \\
A-n33-k6   & 742 & 0.781   & 0.617 & \textbf{0.983} & 0.588 & 0.098  & 0.781                     & {0.817} \\
A-n34-k5   & 778 & 0.475   & 0.397 & \textbf{0.814} & 0.488 & 0.067  & 0.658                     & {0.659} \\
A-n36-k5   & 799 & 0.474   & 0.300 & 0.477          & 0.418 & 0.424  & \textbf{0.704}     & 0.696 \\
A-n37-k5   & 669 & 0.260   & 0.244 & 0.694          & \textbf{0.698} & 0.520              & 0.173      & 0.580 \\
A-n37-k6   & 949 & 0.639   & 0.419 & 0.784          & 0.287 & 0.253  & 0.755                      & \textbf{0.800} \\
A-n38-k5   & 730 & 0.344   & 0.368 & 0.514          & 0.518 & 0.286  & \textbf{0.744}      & 0.711 \\
A-n39-k5   & 822 & 0.590   & 0.378 & 0.348          & 0.220 & 0.250  & \textbf{0.747}        & 0.657 \\
A-n39-k6   & 831 & 0.567   & 0.170 & \textbf{0.804}          & 0.507 & 0.469  & 0.761                      & 0.793 \\
A-n44-k6   & 937 & 0.291   & 0.191 & 0.573          & 0.059 & 0.341  & \textbf{0.811}       & 0.750 \\
A-n45-k6   & 944 & 0.484   & 0.190 & \textbf{0.762}          & 0.451 & 0.287  & 0.683                      & 0.706 \\
A-n45-k7   & 1146 & 0.580  & 0.262 & 0.719          & 0.427 & 0.458  & 0.731                     & \textbf{0.886} \\
A-n46-k7   & 914 & 0.350   & 0.224 & 0.768          & 0.078 & 0.287  & 0.724                      & \textbf{0.784} \\
A-n48-k7   & 1073 & 0.290  & 0.293 & 0.725          & 0.289 & 0.204  & 0.284                     & \textbf{0.787} \\
A-n53-k7   & 1010 & 0.167  & 0.238 & 0.523          & 0.464 & 0.036  & 0.705                     & \textbf{0.740} \\
A-n54-k7   & 1167 & 0.439  & 0.222 & 0.675          & 0.350 & 0.361  & 0.707                     & \textbf{0.711} \\
A-n55-k9   & 1073 & 0.233  & 0.133 & 0.593          & 0.264 & 0.063  & 0.657                     & \textbf{0.762} \\
A-n60-k9   & 1354 & 0.388  & 0.361 & 0.612          & 0.074 & 0.371  & 0.663                     & \textbf{0.758} \\
A-n61-k9   & 1034 & 0.160  & 0.208 & 0.582          & 0.363 & 0.370  & 0.427                     & \textbf{0.718} \\
A-n62-k8   & 1288 & 0.225  & 0.217 & 0.611          & 0.412 & 0.068  & 0.648                     & \textbf{0.749} \\
A-n63-k9   & 1616 & 0.149  & 0.134 & 0.572          & 0.265 & 0.395  & 0.708                     & \textbf{0.775} \\
A-n63-k10  & 1314 & 0.479 & 0.314 & 0.647          & 0.000 & 0.061  & 0.767                     & \textbf{0.825} \\
A-n64-k9   & 1401 & 0.424  & 0.361 & 0.667          & 0.343 & 0.054  & 0.732                     & \textbf{0.802} \\
A-n65-k9   & 1174 & 0.318  & 0.145 & 0.478          & 0.188 & 0.080  & 0.598                     & \textbf{0.754} \\
A-n69-k9   & 1159 & 0.148  & 0.192 & 0.530          & 0.113 & 0.164  & 0.429                     & \textbf{0.778} \\
A-n80-k10 & 1763 & 0.164  & 0.159 & 0.517          & 0.065 & 0.082  & 0.736                       & \textbf{0.765} \\ \midrule
Time $\downarrow$&- &3.0S     & 3.0S    & 3.5M           & 14.0M & 48.0M  & 9.1H                            & 8.1H \\ 
\midrule\bottomrule
\end{tabular}%
}\label{CVRPLIBAPP}
\end{table}

\subsection{Extended Experiments}
\label{CVRPdetails}
\smallskip\noindent\textbf{CVRP.} This section presents more experimental details on the CVRPLIB. 
Taking A-n32-k5 as an example, n32 represents the total 32 nodes including the depot, and k5 represents five vehicles. The cost with respects to the optimal solution is also included.
As observed, POMO~\cite{kwon2020pomo}, Sym-NCO~\cite{kimsym}, and MDAM~\cite{xin2021multi} exhibit highly unstable MSQI, while LEHD~\cite{NEURIPS2023_1c10d0c0}, EAS~\cite{hottung2022efficient} and RF-MA3S perform relatively more consistent with desirable MSQI across instances of different scales.
When the number of nodes is small (i.e., less than 45), the performance of LEHD, EAS and RF-MA3S is comparable. As the number of nodes continues to grow, RF-MA3S consistently outperforms all other methods. Overall, our proposed RF-MA3S presents strong generalization abilities and shows obvious superiority over the compared algorithms in terms of MSQI. Nevertheless, like EAS, our RF-MA3S may also suffer from prolonged running time, leaving room for further optimization.

\begin{table}[htb]
\setlength{\tabcolsep}{4pt}
\centering
\caption{Single-Solution Performance.}
\resizebox{.475\textwidth}{!}{%
\begin{tabular}{@{}l|c|ccc}
\toprule \midrule 
 Method & Model Type & Obj. $\downarrow$ & Gap $\downarrow$ & Time $\downarrow$ \\ \midrule 
Concorde & Exact & 7.765 & - &34.0M\\
\hline 
NeuOpt (D2A=1,T=10k) & L2S/RL &7.766 & $0.02 \%$ &1.0H\\
NeuOpt (D2A=5,T=5k) & L2S/RL &7.765 & $0.00 \%$ &2.1H\\  \hline 
Sym-NCO (A=8,T=200) & L2C/RL & 7.771 & $0.08 \%$ &3.1H\\
POMO (A=8,T=200) & L2C/RL &7.770 & $0.07 \%$ &3.1H\\
POMO+EAS (A=8,T=200) & L2C/RL &7.769 & $0.05 \%$ &6.1H\\
POMO+EAS+SGBS (long) & L2C/RL &7.767 & $0.03 \%$ &0.6D\\ 
LEHD (greedy,T=150) & L2C/RL &7.810 & $0.58 \%$ &0.5M\\
LEHD (RRC=50,T=150) & L2C/RL &7.766 & $0.02 \%$ &7.1M\\
LEHD (RRC=500,T=150) & L2C/RL &7.765 & $0.00 \%$ &1.4H\\ \hline 
RF-MA3S (100)  & L2C/RL & $\mathbf{7.765}$ & $\mathbf{0.00 \%}$ &0.6D\\
\midrule \bottomrule
\end{tabular}}
\label{T13}
\end{table}

\smallskip\noindent\textbf{Single-Solution.} 
Although our RF-MA3S is designed to address multi-solution problems and our proposed MSQI already offers a comprehensive score that accounts for both optimality and diversity, we still compare the performance of RF-MA3S in single-solution scenarios with methods specialized for these tasks, to further elucidate the optimality of our method. We tested RF-MA3S using the TSP100 instances from~\cite{NEURIPS2023_9bae70d3}, and we set the maximum iteration limit of AAS to 200. The results are depicted in Table~\ref{T13}, where L2S and L2C denote the Learning-to-Search and Learning-to-Construct methods, respectively. It is observed that the RF-MA3S also excels in single-solution scenarios, with a performance gap of 0.00\% on the test set of TSP100.

\smallskip\noindent\textbf{Real-World Application.} 
We also apply nine city-scale datasets, each containing 60 real-world POIs, to further evaluate the practical availability of our method. To assess the MSQI, the $L(\pi_{best})$ in Eq.~(\ref{eq:best}) is obtained through OR-Tool~\cite{ortools}. 
 As depicted in Table~\ref{MSQI_REAL}, RF-MA3S consistently delivers superior solutions to other methods across different cities, resulting in excellent MSQI performance. 

\begin{table}[htb]
\setlength{\tabcolsep}{4pt}
\centering
\caption{MSQI performance on real-world dataset.}
\resizebox{.475\textwidth}{!}{%
\begin{tabular}{@{}l|cccccc@{}}
\toprule\midrule
\multirow{2}{*}{Dataset} & \multirow{2}{*}{POMO} & Sym  & MDAM  & MDAM & \multirow{2}{*}{EAS} & RF \\ 
 & & -NCO   & -greedy & -bs   & & -MA3S  \\ 
\midrule
Guangzhou & 0.094& 0.251  & 0.000    & 0.352& 0.248& \textbf{0.723}    \\
London    & 0.085& 0.155  & 0.000    & 0.000& 0.388& \textbf{0.737}    \\
New York  & 0.095& 0.159  & 0.000    & 0.335& 0.000& \textbf{0.681}    \\
Paris     & 0.284& 0.584  & 0.264    & 0.327& 0.091& \textbf{0.623}    \\
Rome      & 0.231& 0.633  & 0.000    & 0.440& 0.377& \textbf{0.704}    \\
Singapore & 0.395& 0.655  & 0.494    & 0.634& 0.053& \textbf{0.701}    \\
Sydney    & 0.271& 0.252  & 0.000    & 0.469& 0.428& \textbf{0.479}    \\
Tokyo     & 0.189& 0.404  & 0.000    & 0.638& 0.241& \textbf{0.728}    \\
Vancouver & 0.139& 0.385  & 0.369    & 0.398& 0.283& \textbf{0.771}    \\\midrule
Time $\downarrow$& 7S& 7S & 96S & 10.9M     & 132M  & 130M \\ 
\midrule\bottomrule
\end{tabular}
}\label{MSQI_REAL}
\end{table}

\section{Conclusions and future works}

We introduce a novel method, RF-MA3S, which applies neural optimization techniques to the diversity optimization of TSP. We first propose the Relativization Filter (RF), designed to make the encoder invariant to affine transformations, thereby improving encoding efficiency. Additionally, our model exploits mult-attentive decoders together with an adaptive active search mechanism (MA3S), which effectively leverages the trade-off between exploration and exploitation through dynamically switching the baseline according to the convergence degree of the model towards global and local interests. As such, our model is able to pursue diverse yet high-quality solutions. Through these novel strategies, RF-MA3S not only surpasses other neural heuristic methods in diversity optimization but also demonstrates competitive performance compared to state-of-the-art traditional heuristics in the field.

Despite the promising results, our approach also has limitations that may open up several interesting future research directions. Firstly, our Relativization Filter (RF) does not perfectly handle mirror transformations, leaving potential for enhancement. Secondly, incorporating additional mechanisms, such as Simulation Guided Beam Search (SGBS)~\cite{choo2022simulation}, Random Re-Construct (RRC)~\cite{NEURIPS2023_1c10d0c0} and so on, could potentially boost the performance. Lastly, given the demonstrated potential of neural approaches for optimizing diversity, it would be worthwhile to explore other promising models and training strategies to further improve the performance, and advance this field as well.

\section{Acknowledgement}

This work was supported in part by the Guangdong Provincial Natural Science Foundation for Outstanding Youth Team Project (Grant No. 2024B1515040010), in part by the National Research Foundation, Singapore, under its AI Singapore Programme (AISG Award No. AISG3-RP-2022-031), in part by the National Natural Science Foundation of China (Grant No. 62276100), and in part by the Guangdong Natural Science Funds for Distinguished Young Scholars (Grant No. 2022B1515020049).

\newpage
\bibliographystyle{ACM-Reference-Format}
\bibliography{KDD_ref}


\begin{thebibliography}{44}


\ifx \showCODEN    \undefined \def \showCODEN     #1{\unskip}     \fi
\ifx \showDOI      \undefined \def \showDOI       #1{#1}\fi
\ifx \showISBNx    \undefined \def \showISBNx     #1{\unskip}     \fi
\ifx \showISBNxiii \undefined \def \showISBNxiii  #1{\unskip}     \fi
\ifx \showISSN     \undefined \def \showISSN      #1{\unskip}     \fi
\ifx \showLCCN     \undefined \def \showLCCN      #1{\unskip}     \fi
\ifx \shownote     \undefined \def \shownote      #1{#1}          \fi
\ifx \showarticletitle \undefined \def \showarticletitle #1{#1}   \fi
\ifx \showURL      \undefined \def \showURL       {\relax}        \fi
\providecommand\bibfield[2]{#2}
\providecommand\bibinfo[2]{#2}
\providecommand\natexlab[1]{#1}
\providecommand\showeprint[2][]{arXiv:#2}

\bibitem[Advani et~al\mbox{.}(2023)]%
        {10.1145/3580305.3599425}
\bibfield{author}{\bibinfo{person}{Rishi Advani}, \bibinfo{person}{Paolo Papotti}, {and} \bibinfo{person}{Abolfazl Asudeh}.} \bibinfo{year}{2023}\natexlab{}.
\newblock \showarticletitle{Maximizing Neutrality in News Ordering}. In \bibinfo{booktitle}{\emph{Proceedings of the 29th ACM SIGKDD Conference on Knowledge Discovery and Data Mining}} (Long Beach, CA, USA) \emph{(\bibinfo{series}{KDD '23})}. \bibinfo{publisher}{Association for Computing Machinery}, \bibinfo{address}{New York, NY, USA}, \bibinfo{pages}{11–24}.
\newblock
\showISBNx{9798400701030}


\bibitem[Angus(2006)]%
        {angus2006niching}
\bibfield{author}{\bibinfo{person}{Daniel Angus}.} \bibinfo{year}{2006}\natexlab{}.
\newblock \showarticletitle{Niching for population-based ant colony optimization}. In \bibinfo{booktitle}{\emph{2006 Second IEEE International Conference on e-Science and Grid Computing (e-Science'06)}}. IEEE, \bibinfo{publisher}{IEEE Computer Society}, \bibinfo{address}{USA}, \bibinfo{pages}{115--115}.
\newblock


\bibitem[Augerat(1995)]%
        {CVRPLIBdata}
\bibfield{author}{\bibinfo{person}{P. Augerat}.} \bibinfo{year}{1995}\natexlab{}.
\newblock \bibinfo{title}{{Set A}}.
\newblock
\newblock
\newblock
\shownote{http://vrp.atd-lab.inf.puc-rio.br/index.php/en/}.


\bibitem[Bahdanau et~al\mbox{.}(2014)]%
        {bahdanau2014neural}
\bibfield{author}{\bibinfo{person}{Dzmitry Bahdanau}, \bibinfo{person}{Kyunghyun Cho}, {and} \bibinfo{person}{Yoshua Bengio}.} \bibinfo{year}{2014}\natexlab{}.
\newblock \showarticletitle{Neural machine translation by jointly learning to align and translate}.
\newblock \bibinfo{journal}{\emph{arXiv preprint arXiv:1409.0473}} (\bibinfo{year}{2014}).
\newblock


\bibitem[Bello et~al\mbox{.}(2017)]%
        {bello2016neural}
\bibfield{author}{\bibinfo{person}{Irwan Bello}, \bibinfo{person}{Hieu Pham}, \bibinfo{person}{Quoc~V Le}, \bibinfo{person}{Mohammad Norouzi}, {and} \bibinfo{person}{Samy Bengio}.} \bibinfo{year}{2017}\natexlab{}.
\newblock \showarticletitle{Neural combinatorial optimization with reinforcement learning}, In \bibinfo{booktitle}{International Conference on Learning Representations}.
\newblock \bibinfo{journal}{\emph{arXiv preprint arXiv:1611.09940}}.
\newblock


\bibitem[Bengio et~al\mbox{.}(2021)]%
        {bengio2021machine}
\bibfield{author}{\bibinfo{person}{Yoshua Bengio}, \bibinfo{person}{Andrea Lodi}, {and} \bibinfo{person}{Antoine Prouvost}.} \bibinfo{year}{2021}\natexlab{}.
\newblock \showarticletitle{Machine learning for combinatorial optimization: a methodological tour d’horizon}.
\newblock \bibinfo{journal}{\emph{European Journal of Operational Research}} \bibinfo{volume}{290}, \bibinfo{number}{2} (\bibinfo{year}{2021}), \bibinfo{pages}{405--421}.
\newblock


\bibitem[Choo et~al\mbox{.}(2022)]%
        {choo2022simulation}
\bibfield{author}{\bibinfo{person}{Jinho Choo}, \bibinfo{person}{Yeong-Dae Kwon}, \bibinfo{person}{Jihoon Kim}, \bibinfo{person}{Jeongwoo Jae}, \bibinfo{person}{Andr\'{e} Hottung}, \bibinfo{person}{Kevin Tierney}, {and} \bibinfo{person}{Youngjune Gwon}.} \bibinfo{year}{2022}\natexlab{}.
\newblock \showarticletitle{Simulation-guided Beam Search for Neural Combinatorial Optimization}. In \bibinfo{booktitle}{\emph{Advances in Neural Information Processing Systems}}, Vol.~\bibinfo{volume}{35}. \bibinfo{publisher}{Curran Associates, Inc.}, \bibinfo{pages}{8760--8772}.
\newblock


\bibitem[Dhulipala et~al\mbox{.}(2016)]%
        {10.1145/2939672.2939862}
\bibfield{author}{\bibinfo{person}{Laxman Dhulipala}, \bibinfo{person}{Igor Kabiljo}, \bibinfo{person}{Brian Karrer}, \bibinfo{person}{Giuseppe Ottaviano}, \bibinfo{person}{Sergey Pupyrev}, {and} \bibinfo{person}{Alon Shalita}.} \bibinfo{year}{2016}\natexlab{}.
\newblock \showarticletitle{Compressing Graphs and Indexes with Recursive Graph Bisection}. In \bibinfo{booktitle}{\emph{Proceedings of the 22nd ACM SIGKDD International Conference on Knowledge Discovery and Data Mining}} (San Francisco, California, USA) \emph{(\bibinfo{series}{KDD '16})}. \bibinfo{publisher}{Association for Computing Machinery}, \bibinfo{address}{New York, NY, USA}, \bibinfo{pages}{1535–1544}.
\newblock
\showISBNx{9781450342322}


\bibitem[Do et~al\mbox{.}(2022)]%
        {do2022niching}
\bibfield{author}{\bibinfo{person}{Anh~Viet Do}, \bibinfo{person}{Mingyu Guo}, \bibinfo{person}{Aneta Neumann}, {and} \bibinfo{person}{Frank Neumann}.} \bibinfo{year}{2022}\natexlab{}.
\newblock \showarticletitle{Niching-based evolutionary diversity optimization for the traveling salesperson problem}. In \bibinfo{booktitle}{\emph{Proceedings of the Genetic and Evolutionary Computation Conference}}. \bibinfo{publisher}{Association for Computing Machinery}, \bibinfo{address}{New York, NY, USA}, \bibinfo{pages}{684–693}.
\newblock


\bibitem[Duan et~al\mbox{.}(2020)]%
        {10.1145/3394486.3403355}
\bibfield{author}{\bibinfo{person}{Lu Duan}, \bibinfo{person}{Haoyuan Hu}, \bibinfo{person}{Zili Wu}, \bibinfo{person}{Guozheng Li}, \bibinfo{person}{Xinhang Zhang}, \bibinfo{person}{Yu Gong}, {and} \bibinfo{person}{Yinghui Xu}.} \bibinfo{year}{2020}\natexlab{}.
\newblock \showarticletitle{Balanced Order Batching with Task-Oriented Graph Clustering}. In \bibinfo{booktitle}{\emph{Proceedings of the 26th ACM SIGKDD International Conference on Knowledge Discovery \& Data Mining}} (Virtual Event, CA, USA) \emph{(\bibinfo{series}{KDD '20})}. \bibinfo{publisher}{Association for Computing Machinery}, \bibinfo{address}{New York, NY, USA}, \bibinfo{pages}{3044–3053}.
\newblock
\showISBNx{9781450379984}


\bibitem[Est{\'e}vez et~al\mbox{.}(2009)]%
        {estevez2009normalized}
\bibfield{author}{\bibinfo{person}{Pablo~A Est{\'e}vez}, \bibinfo{person}{Michel Tesmer}, \bibinfo{person}{Claudio~A Perez}, {and} \bibinfo{person}{Jacek~M Zurada}.} \bibinfo{year}{2009}\natexlab{}.
\newblock \showarticletitle{Normalized mutual information feature selection}.
\newblock \bibinfo{journal}{\emph{IEEE Transactions on neural networks}} \bibinfo{volume}{20}, \bibinfo{number}{2} (\bibinfo{year}{2009}), \bibinfo{pages}{189--201}.
\newblock


\bibitem[Grinsztajn et~al\mbox{.}(2022)]%
        {grinsztajn2022population}
\bibfield{author}{\bibinfo{person}{Nathan Grinsztajn}, \bibinfo{person}{Daniel Furelos-Blanco}, {and} \bibinfo{person}{Thomas~D Barrett}.} \bibinfo{year}{2022}\natexlab{}.
\newblock \showarticletitle{Population-Based Reinforcement Learning for Combinatorial Optimization}.
\newblock \bibinfo{journal}{\emph{arXiv preprint arXiv:2210.03475}} (\bibinfo{year}{2022}).
\newblock


\bibitem[Han et~al\mbox{.}(2018)]%
        {han2018multimodal}
\bibfield{author}{\bibinfo{person}{Xin-Chi Han}, \bibinfo{person}{Hao-Wen Ke}, \bibinfo{person}{Yue-Jiao Gong}, \bibinfo{person}{Ying Lin}, \bibinfo{person}{Wei-Li Liu}, {and} \bibinfo{person}{Jun Zhang}.} \bibinfo{year}{2018}\natexlab{}.
\newblock \showarticletitle{Multimodal optimization of traveling salesman problem: a niching ant colony system}. In \bibinfo{booktitle}{\emph{Proceedings of the Genetic and Evolutionary Computation Conference Companion}}. \bibinfo{publisher}{Association for Computing Machinery}, \bibinfo{address}{New York, NY, USA}, \bibinfo{pages}{87–88}.
\newblock


\bibitem[Hottung et~al\mbox{.}(2022)]%
        {hottung2022efficient}
\bibfield{author}{\bibinfo{person}{Andr{\'e} Hottung}, \bibinfo{person}{Yeong-Dae Kwon}, {and} \bibinfo{person}{Kevin Tierney}.} \bibinfo{year}{2022}\natexlab{}.
\newblock \showarticletitle{Efficient Active Search for Combinatorial Optimization Problems}. In \bibinfo{booktitle}{\emph{International Conference on Learning Representations}}.
\newblock


\bibitem[Huang et~al\mbox{.}(2019)]%
        {huang2019niching}
\bibfield{author}{\bibinfo{person}{Ting Huang}, \bibinfo{person}{Yue-Jiao Gong}, \bibinfo{person}{Sam Kwong}, \bibinfo{person}{Hua Wang}, {and} \bibinfo{person}{Jun Zhang}.} \bibinfo{year}{2019}\natexlab{}.
\newblock \showarticletitle{A niching memetic algorithm for multi-solution traveling salesman problem}.
\newblock \bibinfo{journal}{\emph{IEEE Transactions on Evolutionary Computation}} \bibinfo{volume}{24}, \bibinfo{number}{3} (\bibinfo{year}{2019}), \bibinfo{pages}{508--522}.
\newblock


\bibitem[Huang et~al\mbox{.}(2018)]%
        {huang2018seeking}
\bibfield{author}{\bibinfo{person}{Ting Huang}, \bibinfo{person}{Yue-Jiao Gong}, {and} \bibinfo{person}{Jun Zhang}.} \bibinfo{year}{2018}\natexlab{}.
\newblock \showarticletitle{Seeking multiple solutions of combinatorial optimization problems: A proof of principle study}. In \bibinfo{booktitle}{\emph{2018 IEEE Symposium Series on Computational Intelligence (SSCI)}}. IEEE, \bibinfo{pages}{1212--1218}.
\newblock


\bibitem[Hussain et~al\mbox{.}(2022)]%
        {10.1145/3534678.3539296}
\bibfield{author}{\bibinfo{person}{Md~Shamim Hussain}, \bibinfo{person}{Mohammed~J. Zaki}, {and} \bibinfo{person}{Dharmashankar Subramanian}.} \bibinfo{year}{2022}\natexlab{}.
\newblock \showarticletitle{Global Self-Attention as a Replacement for Graph Convolution}. In \bibinfo{booktitle}{\emph{Proceedings of the 28th ACM SIGKDD Conference on Knowledge Discovery and Data Mining}} (Washington DC, USA) \emph{(\bibinfo{series}{KDD '22})}. \bibinfo{publisher}{Association for Computing Machinery}, \bibinfo{address}{New York, NY, USA}, \bibinfo{pages}{655–665}.
\newblock
\showISBNx{9781450393850}


\bibitem[Joshi et~al\mbox{.}(2019)]%
        {joshi2019efficient}
\bibfield{author}{\bibinfo{person}{Chaitanya~K Joshi}, \bibinfo{person}{Thomas Laurent}, {and} \bibinfo{person}{Xavier Bresson}.} \bibinfo{year}{2019}\natexlab{}.
\newblock \showarticletitle{An efficient graph convolutional network technique for the travelling salesman problem}.
\newblock \bibinfo{journal}{\emph{arXiv preprint arXiv:1906.01227}} (\bibinfo{year}{2019}).
\newblock


\bibitem[Kim et~al\mbox{.}(2021)]%
        {kim2021learning}
\bibfield{author}{\bibinfo{person}{Minsu Kim}, \bibinfo{person}{Jinkyoo Park}, {et~al\mbox{.}}} \bibinfo{year}{2021}\natexlab{}.
\newblock \showarticletitle{Learning collaborative policies to solve NP-hard routing problems}. In \bibinfo{booktitle}{\emph{Advances in Neural Information Processing Systems}}, Vol.~\bibinfo{volume}{34}. \bibinfo{pages}{10418--10430}.
\newblock


\bibitem[Kim et~al\mbox{.}(2022)]%
        {kimsym}
\bibfield{author}{\bibinfo{person}{Minsu Kim}, \bibinfo{person}{Junyoung Park}, {and} \bibinfo{person}{Jinkyoo Park}.} \bibinfo{year}{2022}\natexlab{}.
\newblock \showarticletitle{Sym-NCO: Leveraging Symmetricity for Neural Combinatorial Optimization}. In \bibinfo{booktitle}{\emph{Advances in Neural Information Processing Systems}}, Vol.~\bibinfo{volume}{35}. \bibinfo{publisher}{Curran Associates, Inc.}, \bibinfo{pages}{1936--1949}.
\newblock


\bibitem[Konda and Tsitsiklis(1999)]%
        {konda1999actor}
\bibfield{author}{\bibinfo{person}{Vijay Konda} {and} \bibinfo{person}{John Tsitsiklis}.} \bibinfo{year}{1999}\natexlab{}.
\newblock \showarticletitle{Actor-Critic Algorithms}. In \bibinfo{booktitle}{\emph{Advances in Neural Information Processing Systems}}, Vol.~\bibinfo{volume}{12}. \bibinfo{publisher}{MIT Press}.
\newblock


\bibitem[Kool et~al\mbox{.}(2019)]%
        {kool2018attention}
\bibfield{author}{\bibinfo{person}{Wouter Kool}, \bibinfo{person}{Herke van Hoof}, {and} \bibinfo{person}{Max Welling}.} \bibinfo{year}{2019}\natexlab{}.
\newblock \showarticletitle{Attention, Learn to Solve Routing Problems!}. In \bibinfo{booktitle}{\emph{International Conference on Learning Representations}}.
\newblock


\bibitem[Kwon et~al\mbox{.}(2020)]%
        {kwon2020pomo}
\bibfield{author}{\bibinfo{person}{Yeong-Dae Kwon}, \bibinfo{person}{Jinho Choo}, \bibinfo{person}{Byoungjip Kim}, \bibinfo{person}{Iljoo Yoon}, \bibinfo{person}{Youngjune Gwon}, {and} \bibinfo{person}{Seungjai Min}.} \bibinfo{year}{2020}\natexlab{}.
\newblock \showarticletitle{Pomo: Policy optimization with multiple optima for reinforcement learning}.
\newblock \bibinfo{journal}{\emph{Advances in Neural Information Processing Systems}}  \bibinfo{volume}{33} (\bibinfo{year}{2020}), \bibinfo{pages}{21188--21198}.
\newblock


\bibitem[Li et~al\mbox{.}(2013)]%
        {li2013benchmark}
\bibfield{author}{\bibinfo{person}{Xiaodong Li}, \bibinfo{person}{Andries Engelbrecht}, {and} \bibinfo{person}{Michael~G Epitropakis}.} \bibinfo{year}{2013}\natexlab{}.
\newblock \showarticletitle{Benchmark functions for CEC’2013 special session and competition on niching methods for multimodal function optimization}.
\newblock \bibinfo{journal}{\emph{RMIT University, Evolutionary Computation and Machine Learning Group, Australia, Tech. Rep}} (\bibinfo{year}{2013}).
\newblock


\bibitem[Liu et~al\mbox{.}(2021)]%
        {liu2021multi}
\bibfield{author}{\bibinfo{person}{Yiping Liu}, \bibinfo{person}{Liting Xu}, \bibinfo{person}{Yuyan Han}, \bibinfo{person}{Naoki Masuyama}, \bibinfo{person}{Yusuke Nojima}, \bibinfo{person}{Hisao Ishibuchi}, {and} \bibinfo{person}{Gary~G Yen}.} \bibinfo{year}{2021}\natexlab{}.
\newblock \showarticletitle{Multi-modal multi-objective traveling salesman problem and its evolutionary optimizer}. In \bibinfo{booktitle}{\emph{2021 IEEE International Conference on Systems, Man, and Cybernetics (SMC)}}. IEEE, \bibinfo{publisher}{IEEE Press}, \bibinfo{pages}{770--777}.
\newblock


\bibitem[Liu et~al\mbox{.}(2023)]%
        {liu2023evolutionary}
\bibfield{author}{\bibinfo{person}{Yiping Liu}, \bibinfo{person}{Liting Xu}, \bibinfo{person}{Yuyan Han}, \bibinfo{person}{Xiangxiang Zeng}, \bibinfo{person}{Gary~G Yen}, {and} \bibinfo{person}{Hisao Ishibuchi}.} \bibinfo{year}{2023}\natexlab{}.
\newblock \showarticletitle{Evolutionary multimodal multiobjective optimization for traveling salesman problems}.
\newblock \bibinfo{journal}{\emph{IEEE Transactions on Evolutionary Computation}} \bibinfo{volume}{28}, \bibinfo{number}{2} (\bibinfo{year}{2023}), \bibinfo{pages}{516--530}.
\newblock


\bibitem[Luo et~al\mbox{.}(2023)]%
        {NEURIPS2023_1c10d0c0}
\bibfield{author}{\bibinfo{person}{Fu Luo}, \bibinfo{person}{Xi Lin}, \bibinfo{person}{Fei Liu}, \bibinfo{person}{Qingfu Zhang}, {and} \bibinfo{person}{Zhenkun Wang}.} \bibinfo{year}{2023}\natexlab{}.
\newblock \showarticletitle{Neural Combinatorial Optimization with Heavy Decoder: Toward Large Scale Generalization}. In \bibinfo{booktitle}{\emph{Advances in Neural Information Processing Systems}}, Vol.~\bibinfo{volume}{36}. \bibinfo{publisher}{Curran Associates, Inc.}, \bibinfo{pages}{8845--8864}.
\newblock


\bibitem[Ma et~al\mbox{.}(2023)]%
        {NEURIPS2023_9bae70d3}
\bibfield{author}{\bibinfo{person}{Yining Ma}, \bibinfo{person}{Zhiguang Cao}, {and} \bibinfo{person}{Yeow~Meng Chee}.} \bibinfo{year}{2023}\natexlab{}.
\newblock \showarticletitle{Learning to Search Feasible and Infeasible Regions of Routing Problems with Flexible Neural k-Opt}. In \bibinfo{booktitle}{\emph{Advances in Neural Information Processing Systems}}, Vol.~\bibinfo{volume}{36}. \bibinfo{publisher}{Curran Associates, Inc.}, \bibinfo{pages}{49555--49578}.
\newblock


\bibitem[Mazyavkina et~al\mbox{.}(2021)]%
        {MAZYAVKINA2021105400}
\bibfield{author}{\bibinfo{person}{Nina Mazyavkina}, \bibinfo{person}{Sergey Sviridov}, \bibinfo{person}{Sergei Ivanov}, {and} \bibinfo{person}{Evgeny Burnaev}.} \bibinfo{year}{2021}\natexlab{}.
\newblock \showarticletitle{Reinforcement learning for combinatorial optimization: A survey}.
\newblock \bibinfo{journal}{\emph{Computers \& Operations Research}}  \bibinfo{volume}{134} (\bibinfo{year}{2021}), \bibinfo{pages}{105400}.
\newblock


\bibitem[Perron and Furnon({[n.\,d.]})]%
        {ortools}
\bibfield{author}{\bibinfo{person}{Laurent Perron} {and} \bibinfo{person}{Vincent Furnon}.} \bibinfo{year}{[n.\,d.]}\natexlab{}.
\newblock \bibinfo{booktitle}{\emph{OR-Tools}}.
\newblock Google.
\newblock
\urldef\tempurl%
\url{https://developers.google.com/optimization/}
\showURL{%
\tempurl}


\bibitem[Reinelt(1991)]%
        {reinelt1991tsplib}
\bibfield{author}{\bibinfo{person}{Gerhard Reinelt}.} \bibinfo{year}{1991}\natexlab{}.
\newblock \showarticletitle{{TSPLIB-A} traveling salesman problem library}.
\newblock \bibinfo{journal}{\emph{{ORSA} journal on computing}} \bibinfo{volume}{3}, \bibinfo{number}{4} (\bibinfo{year}{1991}), \bibinfo{pages}{376--384}.
\newblock


\bibitem[Ronald(1995)]%
        {ronald1995finding}
\bibfield{author}{\bibinfo{person}{Simon Ronald}.} \bibinfo{year}{1995}\natexlab{}.
\newblock \showarticletitle{Finding multiple solutions with an evolutionary algorithm}. In \bibinfo{booktitle}{\emph{Proceedings of 1995 IEEE International Conference on Evolutionary Computation}}, Vol.~\bibinfo{volume}{2}. IEEE, \bibinfo{pages}{641--646}.
\newblock


\bibitem[Tong et~al\mbox{.}(2021)]%
        {tong2021combinatorial}
\bibfield{author}{\bibinfo{person}{Yongxin Tong}, \bibinfo{person}{Dingyuan Shi}, \bibinfo{person}{Yi Xu}, \bibinfo{person}{Weifeng Lv}, \bibinfo{person}{Zhiwei Qin}, {and} \bibinfo{person}{Xiaocheng Tang}.} \bibinfo{year}{2021}\natexlab{}.
\newblock \showarticletitle{Combinatorial optimization meets reinforcement learning: Effective taxi order dispatching at large-scale}.
\newblock \bibinfo{journal}{\emph{IEEE Transactions on Knowledge and Data Engineering}} \bibinfo{volume}{35}, \bibinfo{number}{10} (\bibinfo{year}{2021}), \bibinfo{pages}{9812--9823}.
\newblock


\bibitem[Vaswani et~al\mbox{.}(2017)]%
        {vaswani2017attention}
\bibfield{author}{\bibinfo{person}{Ashish Vaswani}, \bibinfo{person}{Noam Shazeer}, \bibinfo{person}{Niki Parmar}, \bibinfo{person}{Jakob Uszkoreit}, \bibinfo{person}{Llion Jones}, \bibinfo{person}{Aidan~N Gomez}, \bibinfo{person}{\L~ukasz Kaiser}, {and} \bibinfo{person}{Illia Polosukhin}.} \bibinfo{year}{2017}\natexlab{}.
\newblock \showarticletitle{Attention is All you Need}. In \bibinfo{booktitle}{\emph{Advances in Neural Information Processing Systems}}, Vol.~\bibinfo{volume}{30}. \bibinfo{publisher}{Curran Associates, Inc.}
\newblock


\bibitem[Vinyals et~al\mbox{.}(2015)]%
        {vinyals2015pointer}
\bibfield{author}{\bibinfo{person}{Oriol Vinyals}, \bibinfo{person}{Meire Fortunato}, {and} \bibinfo{person}{Navdeep Jaitly}.} \bibinfo{year}{2015}\natexlab{}.
\newblock \showarticletitle{Pointer Networks}. In \bibinfo{booktitle}{\emph{Advances in Neural Information Processing Systems}}, Vol.~\bibinfo{volume}{28}. \bibinfo{publisher}{Curran Associates, Inc.}
\newblock


\bibitem[Xin et~al\mbox{.}(2020)]%
        {xin2020step}
\bibfield{author}{\bibinfo{person}{Liang Xin}, \bibinfo{person}{Wen Song}, \bibinfo{person}{Zhiguang Cao}, {and} \bibinfo{person}{Jie Zhang}.} \bibinfo{year}{2020}\natexlab{}.
\newblock \showarticletitle{Step-wise deep learning models for solving routing problems}.
\newblock \bibinfo{journal}{\emph{IEEE Transactions on Industrial Informatics}} \bibinfo{volume}{17}, \bibinfo{number}{7} (\bibinfo{year}{2020}), \bibinfo{pages}{4861--4871}.
\newblock


\bibitem[Xin et~al\mbox{.}(2021)]%
        {xin2021multi}
\bibfield{author}{\bibinfo{person}{Liang Xin}, \bibinfo{person}{Wen Song}, \bibinfo{person}{Zhiguang Cao}, {and} \bibinfo{person}{Jie Zhang}.} \bibinfo{year}{2021}\natexlab{}.
\newblock \showarticletitle{Multi-Decoder Attention Model with Embedding Glimpse for Solving Vehicle Routing Problems}. In \bibinfo{booktitle}{\emph{Proceedings of the AAAI Conference on Artificial Intelligence}}, Vol.~\bibinfo{volume}{35}. \bibinfo{pages}{12042--12049}.
\newblock


\bibitem[Ye et~al\mbox{.}(2018)]%
        {10.1145/3219819.3220111}
\bibfield{author}{\bibinfo{person}{Zeyang Ye}, \bibinfo{person}{Lihao Zhang}, \bibinfo{person}{Keli Xiao}, \bibinfo{person}{Wenjun Zhou}, \bibinfo{person}{Yong Ge}, {and} \bibinfo{person}{Yuefan Deng}.} \bibinfo{year}{2018}\natexlab{}.
\newblock \showarticletitle{Multi-User Mobile Sequential Recommendation: An Efficient Parallel Computing Paradigm}. In \bibinfo{booktitle}{\emph{Proceedings of the 24th ACM SIGKDD International Conference on Knowledge Discovery \& Data Mining}} (London, United Kingdom) \emph{(\bibinfo{series}{KDD '18})}. \bibinfo{publisher}{Association for Computing Machinery}, \bibinfo{address}{New York, NY, USA}, \bibinfo{pages}{2624–2633}.
\newblock
\showISBNx{9781450355520}


\bibitem[Yi et~al\mbox{.}(2019)]%
        {yi2019sampling}
\bibfield{author}{\bibinfo{person}{Xinyang Yi}, \bibinfo{person}{Ji Yang}, \bibinfo{person}{Lichan Hong}, \bibinfo{person}{Derek~Zhiyuan Cheng}, \bibinfo{person}{Lukasz Heldt}, \bibinfo{person}{Aditee Kumthekar}, \bibinfo{person}{Zhe Zhao}, \bibinfo{person}{Li Wei}, {and} \bibinfo{person}{Ed Chi}.} \bibinfo{year}{2019}\natexlab{}.
\newblock \showarticletitle{Sampling-bias-corrected neural modeling for large corpus item recommendations}. In \bibinfo{booktitle}{\emph{Proceedings of the 13th ACM Conference on Recommender Systems}}. \bibinfo{publisher}{Association for Computing Machinery}, \bibinfo{address}{New York, NY, USA}, \bibinfo{pages}{269–277}.
\newblock


\bibitem[Yu and Liu(2003)]%
        {yu2003feature}
\bibfield{author}{\bibinfo{person}{Lei Yu} {and} \bibinfo{person}{Huan Liu}.} \bibinfo{year}{2003}\natexlab{}.
\newblock \showarticletitle{Feature selection for high-dimensional data: A fast correlation-based filter solution}. In \bibinfo{booktitle}{\emph{Proceedings of the 20th international conference on machine learning (ICML-03)}}. \bibinfo{pages}{856--863}.
\newblock


\bibitem[Zhang et~al\mbox{.}(2023)]%
        {zhang2023review}
\bibfield{author}{\bibinfo{person}{Cong Zhang}, \bibinfo{person}{Yaoxin Wu}, \bibinfo{person}{Yining Ma}, \bibinfo{person}{Wen Song}, \bibinfo{person}{Zhang Le}, \bibinfo{person}{Zhiguang Cao}, {and} \bibinfo{person}{Jie Zhang}.} \bibinfo{year}{2023}\natexlab{}.
\newblock \showarticletitle{A review on learning to solve combinatorial optimisation problems in manufacturing}.
\newblock \bibinfo{journal}{\emph{IET Collaborative Intelligent Manufacturing}} \bibinfo{volume}{5}, \bibinfo{number}{1} (\bibinfo{year}{2023}), \bibinfo{pages}{e12072}.
\newblock


\bibitem[Zhang et~al\mbox{.}(2022)]%
        {zhang2022learning2}
\bibfield{author}{\bibinfo{person}{Rongkai Zhang}, \bibinfo{person}{Cong Zhang}, \bibinfo{person}{Zhiguang Cao}, \bibinfo{person}{Wen Song}, \bibinfo{person}{Puay~Siew Tan}, \bibinfo{person}{Jie Zhang}, \bibinfo{person}{Bihan Wen}, {and} \bibinfo{person}{Justin Dauwels}.} \bibinfo{year}{2022}\natexlab{}.
\newblock \showarticletitle{Learning to solve multiple-TSP with time window and rejections via deep reinforcement learning}.
\newblock \bibinfo{journal}{\emph{IEEE Transactions on Intelligent Transportation Systems}} \bibinfo{volume}{24}, \bibinfo{number}{1} (\bibinfo{year}{2022}), \bibinfo{pages}{1325--1336}.
\newblock


\bibitem[Zhou et~al\mbox{.}(2023)]%
        {zhou2023learning}
\bibfield{author}{\bibinfo{person}{Jianan Zhou}, \bibinfo{person}{Yaoxin Wu}, \bibinfo{person}{Zhiguang Cao}, \bibinfo{person}{Wen Song}, \bibinfo{person}{Jie Zhang}, {and} \bibinfo{person}{Zhenghua Chen}.} \bibinfo{year}{2023}\natexlab{}.
\newblock \showarticletitle{Learning large neighborhood search for vehicle routing in airport ground handling}.
\newblock \bibinfo{journal}{\emph{IEEE Transactions on Knowledge and Data Engineering}} \bibinfo{volume}{35}, \bibinfo{number}{9} (\bibinfo{year}{2023}), \bibinfo{pages}{9769--9782}.
\newblock


\bibitem[Zong et~al\mbox{.}(2022)]%
        {10.1145/3534678.3539037}
\bibfield{author}{\bibinfo{person}{Zefang Zong}, \bibinfo{person}{Hansen Wang}, \bibinfo{person}{Jingwei Wang}, \bibinfo{person}{Meng Zheng}, {and} \bibinfo{person}{Yong Li}.} \bibinfo{year}{2022}\natexlab{}.
\newblock \showarticletitle{RBG: Hierarchically Solving Large-Scale Routing Problems in Logistic Systems via Reinforcement Learning}. In \bibinfo{booktitle}{\emph{Proceedings of the 28th ACM SIGKDD Conference on Knowledge Discovery and Data Mining}} (Washington DC, USA) \emph{(\bibinfo{series}{KDD '22})}. \bibinfo{publisher}{Association for Computing Machinery}, \bibinfo{address}{New York, NY, USA}, \bibinfo{pages}{4648–4658}.
\newblock
\showISBNx{9781450393850}


\end{thebibliography}

\balance

\newpage

\appendix
\section{Decoder Details} 
After the encoder, we can obtain the graph embedding of the instance, which will be combined with the first and last nodes in the current (partial) route to yield the context node embedding $\hat{h}$. Using the $d_h$-dimensional node embedding $\tilde{h}$ to represent the set of candidate nodes to be selected for route construction, we exemplify the single-head attention calculation in the used multi-head attention mechanism as follows, 
\begin{equation}
q_i=W^Q \hat{h}_i,~k_i=W^K \tilde{h}_i,~v_i=W^V \tilde{h}_i,
\end{equation}
where query $q$, key $k$ and value $v$ have dimensions of $d_k$, $d_k$ and $d_v$, respectively. It is worth noting that $d_k$ and $d_v$ are equal to $\frac{d_h}{M}$, where $M$ signifies the number of heads in multi-head attention mechanism. Moreover, $W^Q \in \mathbb{R}^{d_h\times d_k}$, $W^K \in \mathbb{R}^{d_h\times d_k}$ and $W^V \in \mathbb{R}^{d_h\times d_v}$ are trainable weight matrices. Then the attention score is calculated and normalized as follows,
\begin{equation}
a_{i,j}=
\begin{cases}
\frac{{q_i}^{\mathsf{T}}{k}_j}{\sqrt{d_k}} , \quad &\text{if~} j \neq \pi_{t'}~\forall t'<t, \\
-\infty , \quad & \text{otherwise},
\end{cases}\label{compa1}
\end{equation}
\begin{equation}
a'_{i,j}=\frac{e^{a_{i,j}}}{\sum_{j'} e^{a_{i,j'}}},
\end{equation}
where we follow the conventional designs~\cite{kool2018attention} and mask out already visited nodes before time $t$ (which are invalid for choice).
Afterwards, the attention scores are multiplied with values $v$ to yield a single-head attention output as follows,
\begin{equation}
{h_i'}=\sum_j a'_{i,j}{v}_j.
\end{equation}
The outputs of each single-head attention are then multiplied by the trainable parameter matrix $W^O \in \mathbb{R}^{d_k\times d_h}$ to project the dimensions back to $d_h$ as follows,
\begin{equation}
\breve{h}_i=\sum_m^M W^O_m h_{im}' .
\end{equation}
The above results are then processed through a single-head attention layer, during which infeasible nodes (the ones that have already been visited) are dynamically filtered out. Later, by applying a softmax, the attention scores determine the action distribution over the remaining nodes for route construction. The details are as follows,
\begin{equation}
\hat{q}_i=W^{Q'} \breve{h_i},
\end{equation}
\begin{equation}
\hat{a}_{i,j}=
\begin{cases}
\frac{\hat{q}_i^{\mathsf{T}}{k}_j}{\sqrt{d_k}} , \quad &\text{if~} j \neq \pi_{t'}~\forall t'<t, \\
-\infty , \quad & \text{otherwise},
\end{cases}\label{compa2}
\end{equation}
\begin{equation}
p_{i,j}=p_\theta(\pi_{i,t}=j|s,\pi_{i,1:t-1})=\frac{e^{\hat{a}_{i,j}}}{\sum_{j'} e^{\hat{a}_{i,j'}}}.
\end{equation}
From the distribution, a node is selected either by sampling during training or by greedily selecting the node with the highest probability during inference. This node is then added to the partial solution, continuing until the complete solution, denoted as $\pi$, is derived.

Moreover, during the training process, temperature softmax will be applied to calculate $p$.

\section{Rationale behind MSQI}

\begin{figure}[tbh]
    \centering\includegraphics[width=\linewidth]{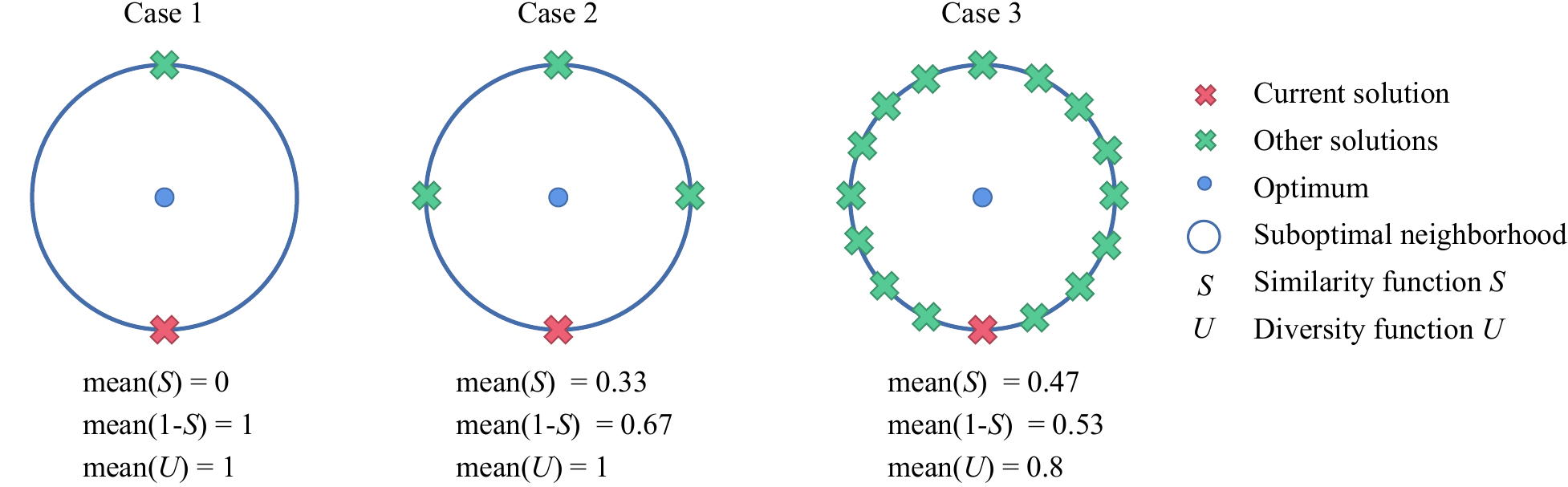}
    \caption{Discussion on the different functions, including $S$, $1-S$ and $U$ used in MSQI.
}
    \label{UaS}
\end{figure}

\begin{figure*} [tbh]
	\centering
	\subfloat[First category.\label{MSTSPLIB:a}]{
		\includegraphics[scale=0.5]{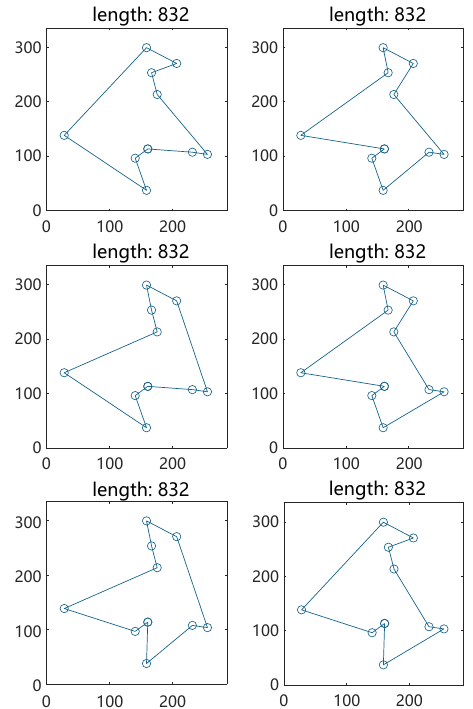}}\quad\quad
	\subfloat[Second category.\label{MSTSPLIB:b}]{
		\includegraphics[scale=0.5]{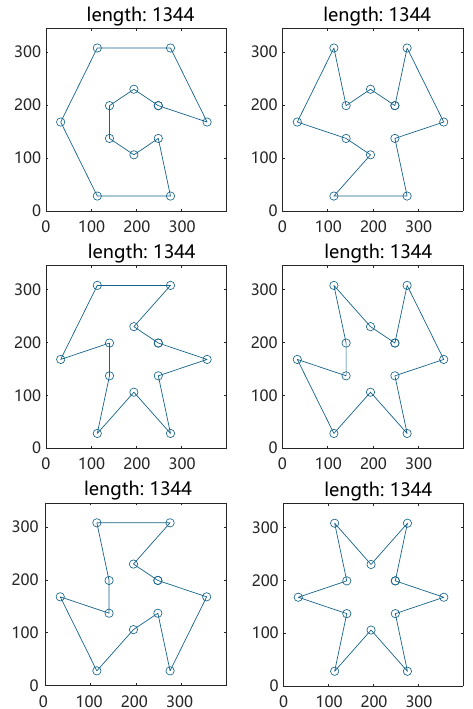}}\quad\quad
	\subfloat[Third category.\label{MSTSPLIB:c}]{
		\includegraphics[scale=0.35]{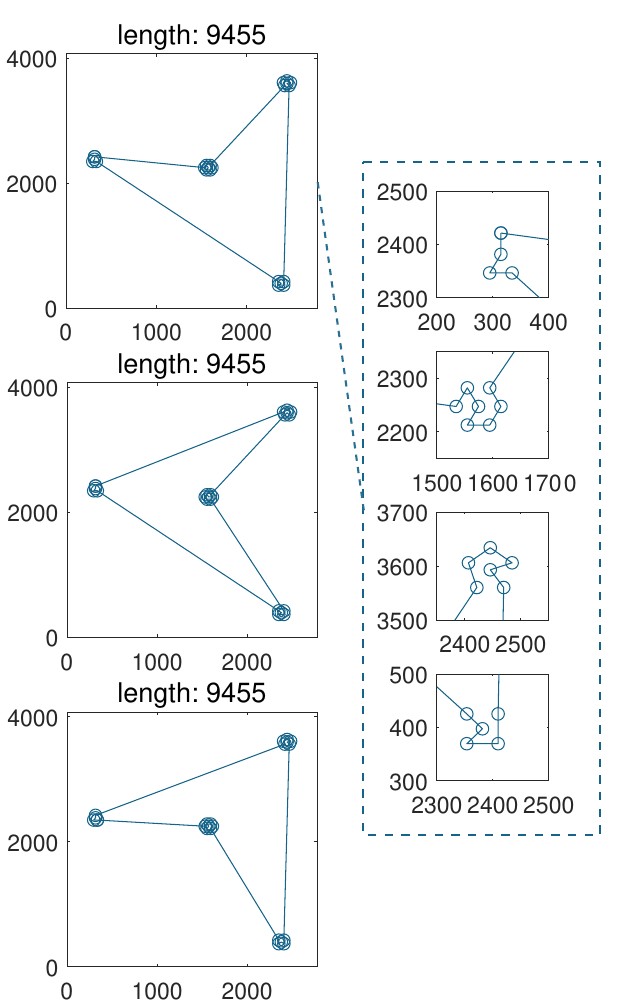}}	\quad\quad
	\subfloat[Fourth category.\label{MSTSPLIB:d}]{
		\includegraphics[scale=0.35]{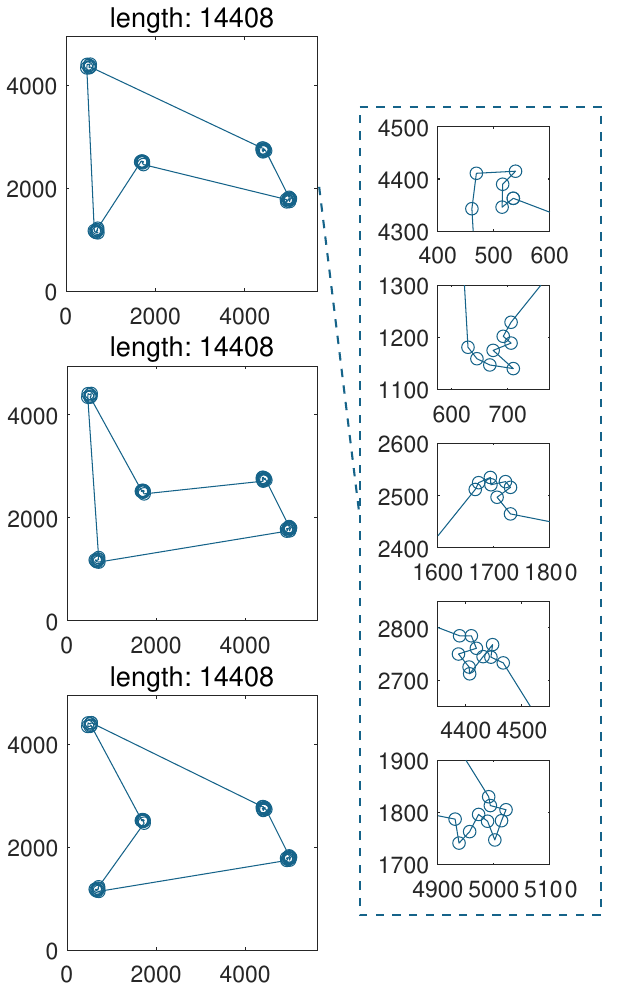} }

	\caption{Overview of MSTSPLIB. }
	\label{fig3} 
\end{figure*}
We discuss more about the rationale behind MSQI where we focus on Eq.~(\ref{fu}). As depicted in Figure ~\ref{UaS}, we illustrate three cases, and compare the $(1-S)$ measure with our refined $U$ measure, using circles to denote suboptimal neighborhoods. In case 1, involving two solutions, a larger distance between them within a given neighborhood signifies a higher degree of difference. The similarity function, denoted as $S$ (i.e., $S$=0 in this case), effectively measures this similarity on a scale of 0 to 1. Consequently, $(1-S)$ quantifies the difference with the same value as our $U$ (i.e., $U$=1). However, a discrepancy arises in case 2. Although the solution distribution displays ideal diversity, $(1-S)$ yields only 0.67, which should be 1 to accurately represent the diversity. To cope with such discrepancy, we introduce the critical value $S=\frac{1}{2}$ and double the output of the equation. In doing so, it ensures that our refined $U$ measure aligns with the anticipated diversity, returning a value of 1 (i.e., $U$=1) within the range of $[0,1]$. In case 3, the presence of more solutions inevitably reduces the distances between them, and thus, $S$ grows up to 0.47. The increase in similarity also correspondingly decreases the diversity function $U$ (i.e., $U$=0.8) which is different from ($1-S$) (i.e., $1-S$=0.53). However, this outcome aligns with our objective of preventing the algorithm from merely increasing solution quantity to boost the diversity measure. Our refined $U$ in MSQI fosters a balance between diversity and optimality, inherently promoting the discovery and identification of representative solutions in diversity optimization.

\subsection{Details on MSTSPLIB}

\label{suite}

 For brevity, the 25 instances in MSTSPLIB are abbreviated as 1-25. 
Detailed information for each instance in MSTSPLIB, including the number of nodes, optimal path length, and the number of ground-truth optimal solutions are presented in Table~\ref{MSTSPLIB dataset}.

\begin{table*}[t]\setlength{\tabcolsep}{1pt}
\caption{MSTSPLIB dataset.}
\centering
\resizebox{.9\textwidth}{!}{%
\begin{tabular}{@{}lccccccccccccccccccccccccc@{}}
\toprule
\midrule
 Category & \multicolumn{6}{|c}{1st} &   \multicolumn{6}{|c}{2nd}  & \multicolumn{4}{|c}{3rd}   & \multicolumn{9}{|c}{4th}\\
Instance & \multicolumn{1}{|c}{1} & 2 & 3 & 4 & 5 & 6 &   \multicolumn{1}{|c}{7} & 8 & 9 & 10 & 11 & 12 &\multicolumn{1}{|c}{13} & 14 & 15 & 16  & \multicolumn{1}{|c}{17} & 18 & 19 & 20 & 21 & 22 & 23 & 24 & 25\\ \midrule
Node No. & \multicolumn{1}{|c}{9} & 10 & 10 & 11 & 12 & 12   & \multicolumn{1}{|c}{10} & 12 & 10 & 10 & 10 & 15 & \multicolumn{1}{|c}{28} & 34 & 22 & 33   & \multicolumn{1}{|c}{35} & 39 & 42 & 45 & 48 & 55 & 59 & 60 & 66\\
Optimum No. & \multicolumn{1}{|c}{3} & 4 & 13 & 4 & 2 & 4  & \multicolumn{1}{|c}{56} & 110 & 4 & 4 & 14 & 196 &\multicolumn{1}{|c}{70} & 16 & 72 & 64   & \multicolumn{1}{|c}{10} & 20 & 20 & 20 & 4 & 9 & 10 & 36 & 26\\
Optimal Cost & \multicolumn{1}{|c}{680} & 1265 & 832 & 803 & 754 & 845   & \multicolumn{1}{|c}{130} & 1344 & 72 & 72 & 78 & 130&\multicolumn{1}{|c}{3055} & 3575 & 9455 & 8761   & \multicolumn{1}{|c}{9061} & 23763 & 14408 & 10973 & 6767 & 10442 & 24451 & 9614 & 9521
  \\ \midrule\bottomrule

\end{tabular}%
}\label{MSTSPLIB dataset}
\end{table*}
\begin{table*}[htb]\setlength{\tabcolsep}{6.9pt}
\caption{The effect of threshold $\alpha$ on optimality and diversity performance.}
\centering
\resizebox{.9\textwidth}{!}{%
\begin{tabular}{@{}l|cc|cc|cc|cc|ccccc@{}}
\toprule\midrule
 & \multicolumn{2}{c|}{1st Category} & \multicolumn{2}{c|}{2nd Category} & \multicolumn{2}{c|}{3rd Category} & \multicolumn{2}{c|}{4th Category} & \multicolumn{5}{c}{Entire test set} \\ 
$\alpha$ & DIFF & OPTI & DIFF & OPTI & DIFF & OPTI & DIFF & OPTI & MSQI & DIFF & OPTI & Solutions & Time \\ \midrule
0.0025 & 0.789 & \textbf{0.810} & 0.808 & \textbf{0.941} & 0.801 & \textbf{0.920} & 0.666 & \textbf{0.923} & 0.754 & 0.751 & \textbf{0.900} & 32.240 & 127.0M \\
0.005 & \textbf{0.811} & 0.783 & \textbf{0.815} & 0.939 & 0.876 & 0.849 & 0.744 & 0.899 & \textbf{0.773} & 0.798 & 0.873 & 80.680 & 40.0M \\
0.01 & \textbf{0.811} & 0.776 & 0.814 & 0.940 & \textbf{0.982} & 0.700 & \textbf{0.857} & 0.832 & 0.755 & \textbf{0.856} & 0.823 & 310.280 & 24.0M \\ \midrule\bottomrule
\end{tabular}%
}\label{alphachange}
\end{table*}

\smallskip\noindent\textbf{First Category.} As shown in Figure~\ref{MSTSPLIB:a}, the first category encompasses instances (mstsp1-6) that are relatively small-scale, containing 9 to 12 randomly distributed nodes.
The ground-truth solution set is derived by brute-force search, and the optima number ranges from 2 to 13. 

\smallskip\noindent\textbf{Second Category.} As shown in Figure~\ref{MSTSPLIB:b}, the second category encompasses instances (mstsp7-12) that are relatively small-scale with symmetric geometric structures, containing 10 to 15 nodes.
The ground-truth solution set is derived inherently during the design of geometries, with a size up to 196. 

\smallskip\noindent\textbf{Third Category.} As shown in Figure~\ref{MSTSPLIB:c}, the third category encompasses instances (mstsp13-16) that are medium-sized composite structures composed of multiple sub-instances with symmetric geometric distributions. These instances contain 22 to 34 nodes, and the number of optimal tours ranges from 16 to 72. 

\smallskip\noindent\textbf{Fourth Category.} As shown in Figure~\ref{MSTSPLIB:d}, the fourth category encompasses instances (mstsp17-25) that consist of large composite structures, comprising multiple randomly distributed sub-instances. These instances contain 35 to 66 nodes, and the number of optimal tours ranges from 4 to 36.
\smallskip\noindent\textbf{First Category.} It encompasses instances (mstsp1-6) that are relatively small-scale, containing 9 to 12 randomly distributed nodes.
The ground-truth solution set is derived by brute-force search, and the optima number ranges from 2 to 13. 

\smallskip\noindent\textbf{Second Category.} It encompasses instances (mstsp7-12) that are relatively small-scale with symmetric geometric structures, containing 10 to 15 nodes.
The ground-truth solution set is derived inherently during the design of geometries, with a size up to 196. 

\smallskip\noindent\textbf{Third Category.} It encompasses instances (mstsp13-16) that are medium-sized composite structures composed of multiple sub-instances with symmetric geometric distributions. These instances contain 22 to 34 nodes, and the number of optimal tours ranges from 16 to 72. 

\smallskip\noindent\textbf{Fourth Category.} It encompasses instances (mstsp17-25) that consist of large composite structures, comprising multiple randomly distributed sub-instances. These instances contain 35 to 66 nodes, and the number of optimal tours ranges from 4 to 36.

\section{Further study}

\smallskip\noindent\textbf{In-Depth Analysis of the $\alpha$ in AAS.} We set $\alpha$ to 0.0025, 0.005, and 0.01 to investigate the effects of parameter variations during training on the MSTSPLIB instances. The model was trained with a dataset of $N=50$, and the maximum number of iterations was set to 250. Table~\ref{alphachange} shows that an excessively large $\alpha$ delays the baseline switching time, and in some cases, the baseline may not switch until reaching the maximum iteration count limit. This prolongs the reliance on the shard baseline, reinforces optimality but compromises diversity, resulting in a decrease in the number of solutions and an increase in iteration time (which could be even longer without a maximum iteration count limit). On the other hand, an excessively small $\alpha$ advances the baseline switching time, ensuring sufficient diversity across solutions but compromising optimality. This leads to a significant increase in the number of solutions within the solution set and reduces the iteration time.

\begin{table}[htb]
\setlength{\tabcolsep}{4pt}
\centering
\caption{MSQI and DI with varying $\delta_1$ in MSTSPLIB.}
\resizebox{.475\textwidth}{!}{%
\begin{tabular}{@{}l|ccc|ccc@{}}
\toprule
\midrule \multirow{2}{*}{ Method } & \multicolumn{3}{c|}{ MSQI $\uparrow$ } & \multicolumn{3}{c}{ DI $\uparrow$ } \\
& 0.1 & 0.01 & 0.001 & 0.1 & 0.01 & 0.001 \\
\midrule NGA & 0.801 & 0.600 & 0.527 & 0.853 & 0.767 & 0.648 \\
NMA & $\mathbf{0 . 8 2 9}$ & $\mathbf{0 . 7 2 5}$ & $\mathbf{0 . 6 2 7}$ & $\mathbf{0 . 9 2 6}$ & $\mathbf{0 . 9 2 6}$ & $\mathbf{0 . 8 3 2}$ \\
\hline POMO (20) & 0.741 & 0.382 & 0.364 & 0.779 & 0.647 & 0.448 \\
Sym-NCO (20) & 0.719 & 0.380 & 0.373 & 0.794 & 0.614 & 0.487 \\
EAS (20) & 0.557 & 0.426 & 0.384 & 0.797 & 0.774 & 0.579 \\
MDAM-greedy (20) & 0.600 & 0.219 & 0.165 & 0.626 & 0.407 & 0.300 \\
MDAM-bs (20) & 0.554 & 0.431 & 0.478 & 0.835 & 0.748 & 0.598 \\
NeuOpt (20) & 0.675 & 0.422 & 0.296 & 0.232 & 0.122 & 0.096 \\
RF-MA3S (20) & $\mathbf{0 . 7 5 5}$ & $\mathbf{0 . 5 3 7}$ & $\mathbf{0 . 4 9 8}$ & $\mathbf{0 . 8 7 7}$ & $\mathbf{0 . 8 7 4}$ & $\mathbf{0 . 7 6 9}$ \\
\hline POMO (50) & 0.621 & 0.253 & 0.188 & 0.676 & 0.459 & 0.324 \\
Sym-NCO (50) & 0.680 & 0.424 & 0.417 & 0.794 & 0.611 & 0.497 \\
EAS (50) & 0.455 & 0.417 & 0.410 & 0.793 & 0.766 & 0.523 \\
MDAM-greedy (50) & 0.381 & 0.063 & 0.046 & 0.491 & 0.121 & 0.071 \\
MDAM-bs (50) & 0.429 & 0.414 & 0.399 & 0.788 & 0.618 & 0.470 \\
NeuOpt (50) & 0.517 & 0.149 & 0.120 & 0.254 & 0.069 & 0.041 \\
RF-MA3S (50) & $\mathbf{0 . 7 3 7}$ & $\mathbf{0 . 4 8 8}$ & $\mathbf{0 . 5 1 0}$ & $\mathbf{0 . 8 8 3}$ & $\mathbf{0 . 8 7 8}$ & $\mathbf{0 . 7 0 5}$ \\
\hline 
LEHD (100) & 0.534 & 0.344 & 0.251 & 0.754 & 0.652 & 0.444 \\
POMO (100) & 0.493 & 0.382 & 0.356 & 0.772 & 0.591 & 0.412 \\
Sym-NCO (100) & 0.531 & 0.264 & 0.201 & 0.738 & 0.507 & 0.322 \\
EAS (100) & 0.389 & 0.421 & 0.412 & 0.742 & 0.620 & 0.496 \\
MDAM-greedy (100) & 0.175 & 0.000 & 0.000 & 0.259 & 0.000 & 0.000 \\
MDAM-bs (100) & 0.256 & 0.252 & 0.216 & 0.558 & 0.353 & 0.275 \\
NeuOpt (100) & 0.608 & 0.258 & 0.159 & 0.276 & 0.128 & 0.092 \\
RF-MA3S (100) & $\mathbf{0 . 6 1 2}$ & $\mathbf{0 . 4 8 0}$ & $\mathbf{0 . 4 2 6}$ & $\mathbf{0 . 8 7 5}$ & $\mathbf{0 . 8 6 7}$ & $\mathbf{0 . 5 3 5}$ \\
\midrule
\bottomrule
\end{tabular}}\label{T12}
\end{table}
\smallskip\noindent\textbf{Analysis of Measures with Varying $\delta_1$ and Excluding $\delta_2$.}
\colorr{Due to the adoption of the harmonic mean in MSQI, it is prone to being influenced by extreme values. Therefore, we choose lenient threshold conditions, which balances the impacts of optimality and diversity on this measure. To further elucidate the merits of our method, we expanded our investigation by adopting a novel perspective. Specifically, we further adjusted the threshold on MSTSPLIB to $\delta_1 = \{0.1, 0.01, 0.001\}$ and $\delta_2 = 0.999$ (removing duplicate solutions only). As shown in Table~\ref{T12}, RF-MA3S exhibits the best performance in both the MSQI and DI compared with other neural heuristic methods.}
\begin{table}[htb]
\setlength{\tabcolsep}{1.8pt}
\centering
\caption{The effect of threshold $\delta$ on MSQI performance.}
\resizebox{.475\textwidth}{!}{%
\begin{tabular}{c|cccccccccc}
\toprule\midrule
\diagbox{$\delta_1$}{$\delta_2$} & 0.1 & 0.2 & 0.3 & 0.4 & 0.5 & 0.6 & 0.7 & 0.8 & 0.9 & 1.0 \\ \midrule
0.1 & 0.438 & 0.466 & 0.475 & 0.578 & 0.741 & 0.848 & 0.811 & 0.773 & 0.760 & 0.749 \\
0.2 & 0.438 & 0.474 & 0.477 & 0.590 & 0.754 & 0.857 & 0.851 & 0.837 & 0.820 & 0.808 \\ \midrule\bottomrule
\end{tabular}%
}\label{deltachange}
\end{table}
\smallskip\noindent\textbf{Impact of the $\delta_1$ and $\delta_2$ in MSQI.} We set $\delta_1=\{0.1,0.2\}$ and $\delta_2=\{0.1,0.2,\cdots,1.0\}$ respectively to examine the impact of threshold settings on MSTSP performance of RF-MA3S in MSTSPLIB. Table~\ref{deltachange} shows that increasing the optimality threshold $\delta_1$ raises the value of the optimality measure, thereby improving the value of MSQI. On the other hand, as the similarity threshold $\delta_2$ increases from small to large, the MSQI measure exhibits an initial increase (in the range of 0.1-0.5) followed by a decrease (in the range of 0.7-1.0). This is because excessively high similarity filtering can compromise the optimality of the solution set, while excessively low similarity filtering retains a larger number of highly similar solutions, thereby reducing the value of the difference measure. 
In light of them, our thresholds within the decreasing range can ensure fair comparisons.

\begin{figure}[tbh] 
    \centering
    \includegraphics[width=\linewidth]{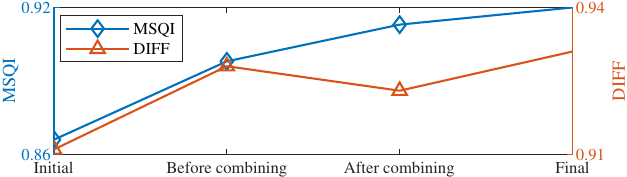}
    \caption{Performance changes pre-post baseline combining.}
    \label{newfig}
\end{figure}

\smallskip\noindent\textbf{Performance changes before and after combining baseline.} \colorb{In the four stages of model inference: the initial solution, before combining baseline, after combining baseline, and the final solution, we recorded the changes(RF-MA3S(50) test on mstsp17) in MSQI and DIFF, as shown in the example in Figure~\ref{newfig}. The MSQI consistently increased, while DIFF temporarily decreased after combining baseline and then increased again. Thus, while a "shared baseline" may reduce diversity, it does not result in either an immediate or significant loss in our method.}

\begin{table}[htb]
\setlength{\tabcolsep}{16pt}
\centering
\caption{Fluctuations in model inference.}
\resizebox{.475\textwidth}{!}{%
\begin{tabular}{@{}l|ccc}
\toprule \midrule
Index & MSQI $\uparrow$  & D $\uparrow$ & Time $\downarrow$ \\ \midrule
Expected Value                & 0.7856                            & 0.8636                          & 24.5500M     \\      
Standard Deviation            & 0.0027                            & 0.0028                          & 2.1761M       \\     
\midrule\bottomrule
\end{tabular}} \label{T15T}
\end{table}

\smallskip\noindent\textbf{Fluctuations in Model Inference.}
\colorb{Regarding the fluctuations of the developed method, we conducted experiments with RF-MA3S(20) on the MSTSPLIB 10 times. As show in the Table~\ref{T15T}, it is evident that RF-MA3S exhibits stable performance.}

\balance

\end{document}